\documentclass[preprint]{elsarticle}
\pdfoutput=1

\usepackage{lineno,hyperref}
\usepackage[utf8]{inputenc}
\modulolinenumbers[5]
\usepackage[fleqn]{amsmath}
\usepackage{amsthm,amssymb,stmaryrd,pifont,wasysym}

\usepackage[toc,page]{appendix} 
\usepackage{lscape}
\usepackage{todonotes}
\usepackage[linesnumbered,lined,boxed,commentsnumbered]{algorithm2e}
\usepackage{xcolor}
\usepackage{graphicx}
\usepackage{caption}
\usepackage{float}

\usepackage[tableposition=top]{caption}
\usepackage{tabularx}
\newcolumntype{L}{>{\raggedright\arraybackslash}X}
\usepackage{multirow}

\makeatletter
\newcommand{\mathleft}{\@fleqntrue\@mathmargin0pt}
\newcommand{\mathcenter}{\@fleqnfalse}
\makeatother

\usepackage{rotating}

\usepackage{lineno,hyperref}
\usepackage{array, longtable, booktabs, makecell}
\usepackage{lipsum}
\usepackage{ragged2e}

\usepackage[T1]{fontenc}
\usepackage[utf8]{inputenc}
\usepackage{pdflscape}
\usepackage{ltablex} 

\usepackage{svg}

\keepXColumns

\usepackage[margin=2.5cm]{geometry}
\usepackage{afterpage}
\usepackage{lscape}
\DeclareMathOperator*{\argmin}{arg\,min}
\DeclareMathOperator*{\argmax}{arg\,max}

\makeatletter
\def\ps@pprintTitle{%
 \let\@oddhead\@empty
 \let\@evenhead\@empty
 \def\@oddfoot{}%
 \let\@evenfoot\@oddfoot}
\makeatother

\journal{}

\newcommand{\comment}[1]{}

\usepackage{dutchcal} 

\usepackage{pifont}

\usepackage{soul}
\definecolor{aqua}{rgb}{0.0, 1.0, 1.0}

\sethlcolor{aqua}

\begin{document}

\setlength{\belowdisplayskip}{5pt} \setlength{\belowdisplayshortskip}{5pt}
\setlength{\abovedisplayskip}{0pt} \setlength{\abovedisplayshortskip}{0pt}

\begin{frontmatter}

\title{Dynamic operator management in meta-heuristics using reinforcement learning: an application to permutation flowshop scheduling problems}

\author[vu]{Maryam Karimi Mamaghan\corref{aaa}}
\ead{m.karimi.mamaghan@vu.nl}

\author[tue]{Mehrdad Mohammadi}
\ead{m.mohammadi1@tue.nl}

\author[vu]{Wout Dullaert}
\ead{w.e.h.dullaert@vu.nl}

\author[unibo1,unibo2]{Daniele Vigo}
\ead{daniele.vigo@unibo.it}

\author[Kedge]{Amir Pirayesh}
\ead{amir.pirayesh@kedgebs.com}

\cortext[aaa]{Corresponding author.}

\address[vu]{Vrije Universiteit Amsterdam, School of Business and Economics, Department of Operations Analytics, the Netherlands}
\address[tue]{Department of Industrial Engineering and Innovation Sciences, Eindhoven University of Technology, Eindhoven 5600MB, the Netherlands}
\address[unibo1]{DEI “G. Marconi”, University of Bologna, Viale del Risorgimento 2, 40136 Bologna, Italy}
\address[unibo2]{CIRI-ICT, University of Bologna, via Quinto Bucci, 336, 47521 Cesena, Italy}
\address[Kedge]{Centre of Excellence in Supply Chain and Transportation (CESIT), KEDGE Business School, Bordeaux, France}

\begin{abstract}

This study develops a framework based on reinforcement learning to dynamically manage a large portfolio of search operators within meta-heuristics. Using the idea of tabu search, the framework allows for continuous adaptation by temporarily excluding less efficient operators and updating the portfolio composition during the search. A Q-learning-based adaptive operator selection mechanism is used to select the most suitable operator from the dynamically updated portfolio at each stage. Unlike traditional approaches, the proposed framework requires no input from the experts regarding the search operators, allowing domain-specific non-experts to effectively use the framework. The performance of the proposed framework is analyzed through an application to the permutation flowshop scheduling problem. The results demonstrate the superior performance of the proposed framework against state-of-the-art algorithms in terms of optimality gap and convergence speed.

\end{abstract}


\end{frontmatter}

\section{Introduction} \label{sec: introduction}

Generally, meta-heuristics may employ only a single or a portfolio of multiple search operators during the search. Using a portfolio of multiple search operators with different characteristics has been shown to improve the exploration and exploitation ability and, consequently, to enhance the overall performance of the meta-heuristics in solving different combinatorial optimization problems (COPs) \citep{dos2014reactive,li2015iterated,di2015experimental,karimi2022machine,karimi2023learning}. From a theoretical perspective, the search space of a COP represents a non-stationary environment, meaning that the performance of different search operators varies depending on the region of the search space being explored. An operator working well in one region might be less effective in another region. Accordingly, incorporating a portfolio of diverse operators into a meta-heuristic is expected to enhance its overall performance \citep{fialho2010adaptive}. 

For every COP, numerous search operators are available in the literature (either variations of the same operator with different configurations or entirely distinct operators), with the possibility of proposing new ones. Since the operators' performance is not pre-determined but rather dependent on the algorithm's performance on specific problems/instances, predicting the operators' performance proves challenging. Even if the most efficient operators could be determined, the order in which these efficient operators should be involved during the search process remains undetermined. Hence, optimizing the performance of a meta-heuristic with multiple operators for solving different problem instances is always challenging \citep{fialho2010adaptive,fialho2010analyzing,li2013adaptive,mosadegh2020stochastic}. We label this problem as \textit{operator management} problem in meta-heuristics, wherein the user should address two questions: \textit{What operators should I include in the portfolio}?, and \textit{How (in which order) should I involve the in-portfolio operators during the search process}? 

To answer these two questions, \textit{operator management} should therefore encompass two phases: \textit{portfolio determination} and \textit{operator selection}. While the former determines which specific operators (among various available ones) are to be included in the portfolio, the latter focuses on how (in which order) in-portfolio operators are to be selected during the search process. One simple way of operator management consists of including all operators in the portfolio and selecting them randomly during the search process. However, the larger the size of the portfolio, the higher the chance of employing inefficient/inappropriate operators \citep{karimi2022machine,karimi2023learning,yin2024adaptive}. 

Portfolio determination can be made either static or dynamic. In static portfolio determination, the size of the portfolio and the choice of in-portfolio operators are predefined before the search process begins and will remain unchanged throughout the entire search process. In contrast, dynamic portfolio determination involves dynamically determining the size and composition of the portfolio at every stage of the search process. To the best of our knowledge, almost all studies in the meta-heuristic literature focus on a static portfolio determination where the decision on the portfolio composition is left to the user to decide based on the literature, domain-specific knowledge, or a comprehensive pre-processing \citep{dos2014reactive,maturana2010autonomous,mosadegh2020stochastic}. There are several issues with these static approaches. First, the identified good operators from the literature may perform poorly when applied to new instances of the problem at hand. Second, expert knowledge is not always available, and even when available, it may not be sufficient and lead us to sub-optimal decisions \citep{bengio2020machine}. Third, pre-processing could be significantly time-consuming depending on the number of available operators and the size and variety of the instances. Fourth, the operators are only pre-tested on a random/limited sample of problem instances, and the ones with the best overall performance are included in the portfolio. This may lead to overlooking operators that may perform well with other instances but are deemed inefficient based on the sample \citep{fialho2010adaptive}. A naive way to overcome these issues is to include all the available operators in the portfolio. The advantage is that all operators are available whenever needed with less reliance on domain-specific knowledge and no need for time-consuming pre-processing. However, including all available operators in the portfolio may lead to poor convergence of the meta-heuristics toward good solutions, as inappropriate/inefficient operators may be selected and employed at certain stages of the search. Therefore, an approach that can leverage all operators while restricting the portfolio to only the best-performing operators at each stage of the search (i.e., dynamically adapting the portfolio to the search space) is expected to alleviate the issue.

Similar to portfolio determination, the operator selection process can be performed either static or dynamic \citep{fialho2010adaptive}. In a static operator selection, the order or the selection probability of the in-portfolio operators is determined a priori and kept fixed throughout the search process. On the contrary, in a dynamic operator selection, the in-portfolio operators are dynamically selected and employed during the search process based on the problem's evolving characteristics or the search space's state. The simplest dynamic selection approach is the random selection of the operators at each stage of the search, regardless of their performance during the search process. In contrast to such a blind selection, one efficient dynamic approach is to consider the search operators' performance in the selection process. This is called Adaptive Operator Selection (AOS) \cite{fialho2010adaptive,karimi2023learning}. Numerous studies have been conducted on AOS, leading to a significant body of work that explores various aspects of this field \citep{pettinger2002controlling,battiti2008reactive,burke2011adaptive,yuan2014empirical,di2015experimental,kheiri2016iterated,li2016two}. However, the application of machine learning for AOS is an emerging field that has gained considerable attention in recent years \citep{dos2014reactive,gunawan2018adopt,mosadegh2020stochastic, durgut2021adaptive,karimi2023learning}. Several studies model AOS as a multi-armed bandit problem, where each operator is treated as one of \textit{K} independent arms, each with an unknown probability of yielding a reward. The optimal strategy for selecting arms is the one that maximizes cumulative reward over time \citep{li2013adaptive,fialho2010analyzing}. These studies primarily focus on historical rewards, neglecting to adequately consider the future returns of the operators. As the selected operators determine the return in the future rather than the past, it is essential to account for future returns in the selection of the operators. In recent years, a new paradigm in AOS has been formed, where reinforcement learning is used to model the AOS problem \citep{ahmadi2018hybrid,mosadegh2020stochastic,durgut2021adaptive,karimi2023learning}. In reinforcement learning, future returns play a crucial role in making decisions and addressing the issues with the previous approaches. Most studies use the classical Q-learning algorithm to model the AOS problem, wherein the operators are considered as actions and selected according to their performance and the current state of the search. Q-learning has been shown to be successful for AOS in solving different COPs \citep{sakurai2012population,handoko2014reinforcement,dos2014reactive,buzdalova2014selecting,silva2016transmission,gunawan2018adopt,ahmadi2018hybrid,mosadegh2020stochastic,zhao2021cooperative,durgut2021adaptive} where some state-of-the-art results have been obtained for permutation flowshop scheduling problem (\citep{karimi2023learning}. When dealing with high-dimensional or continuous state spaces, some studies use deep Q-learning, which allows the Q-value of the operators to be learned using a deep neural network \citep{yin2024adaptive,che2023deep}.

In this paper, we propose a dynamic operator management framework designed to efficiently manage a large set of search operators within meta-heuristics.

\subsection{Contributions of this paper} \label{sec: contributions}

Studies in the literature addressing operator management in meta-heuristics mostly focus on dynamically making operator selection decisions, leaving portfolio determination decisions static \citep{maturana2010autonomous, karimi2022machine}. However, a successful dynamic operator selection that adapts to different instances requires a dynamic approach to portfolio determination, ensuring a broad range of operators are accessible whenever needed, with the portfolio's composition adjusted based on the characteristics of the search space. 

This is the first study that aims to bridge the gap in the existing literature by proposing a framework wherein both \textit{operator selection} and \textit{portfolio determination} decisions are made dynamically. In this framework, by having access to all operators, the portfolio composition (i.e., the size and the in-portfolio operators) is dynamically updated during the search, allowing to temporarily exclude less-efficient operators and only keep useful ones during the search process. This allows the framework to manage a large set of search operators in meta-heuristics. 

Accordingly, the main contributions of this paper to the literature are as follows:

\begin{itemize}
    \item This paper develops a framework to manage a large set of operators in meta-heuristics for solving COPs. This framework is general and easy to implement as it relieves the need for domain-specific knowledge and simply allows the inclusion of all available operators;  
    \item We integrate the dynamic portfolio determination mechanism into Q-learning to address AOS, wherein Q-learning selects the most appropriate operator from the dynamic portfolio at every stage of the search process based on the operators' performance history and the state of the search;
    \item We deploy the idea of tabu search for dynamic portfolio determination to temporarily remove inefficient operators from the portfolio for a fixed number of iterations. This mechanism helps to avoid cycling over inefficient operators, especially over those operators that are highly likely to be selected by Q-learning due to their high-performance history; 
    \item Despite having theoretically extra complexity overhead, we show that the overhead complexity is independent of the instance size, and
    \item For the permutation flowshop scheduling problem (PFSP), an established COP from the literature, we demonstrate that the proposed framework outperforms the state-of-the-art algorithms in terms of optimality gap and convergence speed.
\end{itemize}

\subsection{Application of the proposed framework} \label{sec: application to PFSP}

We explore the application of the proposed operator management framework to a well-known COP, the PFSP, with the makespan minimization objective function. The PFSP is among the most extensively studied problems in the Operational Research community, with numerous practical applications in manufacturing systems \citep{fernandez2014insertion}. The PFSP becomes NP-hard when the number of machines is greater than two \citep{ruiz2007simple}. Consequently, various efficient meta-heuristics have been developed to solve the PFSP \citep{fernandez2017new}. The Iterated Greedy (IG) algorithm has proven to be effective in solving the PFSP \citep{ruiz2007simple,fernandez2019best,karimi2023learning}, although its exploration performance highly depends on the choice of its perturbation operator(s). 

Depending on the size of the PFSP instances, different perturbation operators with different exploration levels can be designed or configured, each exhibiting promising exploration performances in various stages of the search process. The significant advantage of incorporating multiple perturbation operators in the IG algorithm has been demonstrated by \cite{karimi2023learning}. However, recent studies in the literature have focused on the design of IG algorithms with only a single perturbation operator \citep{ruiz2007simple,dubois2017iterated,fernandez2019best}.  In the work by \cite{karimi2023learning}, a static portfolio of only three perturbation operators is considered where the main goal of limiting the number of operators is to reduce the chance of selecting inappropriate operators at different stages of the search process. In addition to requiring extensive pre-processing to identify the three best operators out of available options, this static portfolio eliminates the opportunity to employ potentially effective operators that were excluded in pre-processing. Hence, in this paper, we are motivated to overcome this limitation by taking advantage of the presence of all perturbation operators while limiting the chance of employing inappropriate ones at every stage of the search process. The framework proposed in this paper deploys the idea of tabu search for dynamic portfolio determination and employs Q-learning to perform AOS.

The rest of this paper is organized as follows. Section \ref{sec: General IG for PFSP} provides background on existing works, introducing the PFSP, the original IG algorithm, reinforcement learning concepts, and the Q-learning algorithm. Section \ref{sec: QIG} presents the proposed operator management framework and discusses its complexity. To investigate the performance of the proposed framework, Section \ref{sec: Experimental design} designs a set of computational experiments, and Section \ref{sec: numerical and statistical results} compares the performance of the proposed framework with its variants (i.e., random operator selection and static portfolio) and the state-of-the-art algorithms. Finally, conclusions are presented in Section \ref{sec: conclusion}.

\section{Background} \label{sec: General IG for PFSP}

This section provides an overview of the existing algorithms (i.e., the original IG algorithm and Q-learning), which serve as the foundation for our proposed framework. Additionally, we describe the PFSP, the problem to which we apply our proposed framework.  

\subsection{Permutation flowshop scheduling problem}

The PFSP is a well-known classic optimization problem that involves scheduling a set of $n$ independent jobs $j = \{1, 2, \dots, n\}$ that must be processed on a set of $m$ machines $i = \{1, 2, \dots, m\}$. Each job $j$ has a specified processing time on each machine $i$ denoted by $p_{ij}$. The objective is to find a permutation of the jobs such that a desired criterion is optimized. The desired criterion is typically to minimize the total completion time, known as makespan ($C_{max}$), or other criteria such as total flow time, total tardiness, or the sum of the total earliness and tardiness. In this study, we focus on minimizing makespan.

To calculate the makespan ($C_{max}$), let $\Pi := (\pi_1, \dots, \pi_n) \in \mathfrak{S}(N)$ denote the permutation of jobs, where $\pi_j \in N$ denotes the index of the job appearing at position $j$. The completion time of job $\pi_j$ on machine $i$, $C_{i\pi_j}$, can be determined in a recursive way using Equation (\ref{eqs: completion time3}), where $C_{0\pi_j}=C_{i\pi_0}=0$. The $C_{max}$ is then calculated as Equation (\ref{eqs: cmax}).

{\setlength{\belowdisplayskip}{0pt} \setlength{\belowdisplayshortskip}{0pt}
\setlength{\abovedisplayskip}{0pt} \setlength{\abovedisplayshortskip}{0pt}

\begin{equation} \label{eqs: completion time3}
    C_{i\pi_j} = \max \{C_{i-1,\pi_j},C_{i,\pi_j-1}\} + p_{i,\pi_j}
\end{equation}
}

\begin{equation} \label{eqs: cmax}
    C_{max} = C_{m\pi_n}
\end{equation}

\subsection{Iterated greedy (IG) algorithm} \label{sec: original IG}

The original IG algorithm was first developed for the PFSP by \cite{ruiz2007simple}. The pseudo-code of the IG is given in Algorithm \ref{alg: IG}. The main idea behind IG is to iteratively apply a greedy constructive heuristic to the solution. This constructive heuristic acts as a perturbation mechanism that involves two steps: first, a solution goes through a destruction step wherein $d$ jobs are randomly removed from $\Pi$. This results in a partial sequence of the remaining jobs $\Pi_{\setminus D}$. Then, in the construction step, the removed jobs are inserted back into the partial sequence in a way that optimizes the objective function. 
The main goal of perturbation is to perturb the current solution with the hope of escaping from a local optimum. The perturbation is followed by a local search where the main goal is to improve the current solution. Despite its simplicity, IG has demonstrated state-of-the-art performance for many different PFSP variants \citep{fernandez2017new}. Various extensions have been proposed to IG, either by adding some extra elements to the algorithm (e.g., adding an optional local search applied to the partial sequence obtained after the destruction \citep{dubois2017iterated}) or by proposing approaches to deal with ties, i.e., different positions that provide the same minimum value of $C_{max}$ when inserting a job into its best position \citep{fernandez2014insertion,vasiljevic2015handling}.

\vspace{0.1cm}
\SetAlgoCaptionSeparator{.}
\IncMargin{1em}

\begin{algorithm}[H]
\caption{Pseudo-code of the IG algorithm}
\label{alg: IG}

\SetKwFunction{FMain}{IG}
\SetKwProg{Fn}{Function}{:}{}

\newcommand\mycommfont[1]{\footnotesize\ttfamily\textcolor{blue}{#1}}
\SetCommentSty{mycommfont}


\DontPrintSemicolon

\KwIn{Problem instance, termination criterion, $d$}
\KwOut{$\Pi^* \in \mathfrak{S}(N)$}

\Fn{\FMain{Problem instance, termination criterion, $d$}}{

    $\Pi := \texttt{generateInitialSolution}()$ 
    
    $\Pi := \texttt{applyLocalSearch}(\Pi)$ 
    
    $\Pi^* := \Pi$ 

    \While{\upshape \texttt{terminationCriterion}()}{
        
        \tcp*[l]{\textbf{Perturbation}}
        
        $D := \texttt{destructSolution}(\Pi,d)$ 
        
        $\Pi_{\setminus D} := \Pi \setminus D$ 
        
        
        $\Pi' := \texttt{constructSolution}(\Pi_{\setminus D}, D)$ 

        \tcp*[l]{\textbf{Local search}}
        $\Pi' := \texttt{applyLocalSearch}(\Pi')$

        \tcp*[l]{\textbf{Acceptance}}
        $\Pi := \texttt{acceptSolution}(\Pi', \Pi)$
        
        $\Pi^* := \texttt{updateBestSolution}(\Pi', \Pi^*)$
       
        
    }
    
    \Return $\Pi^*$
}

\end{algorithm}
\DecMargin{1em}
\vspace{0.1cm}

\subsection{Reinforcement learning and Q-learning}

Reinforcement Learning (RL) is a potent computational framework for addressing decision-making and control problems, where an agent learns to achieve a goal through gaining experience (interacting with its environment) \citep{kaelbling1996reinforcement,sutton2018reinforcement}. The fundamental principle of RL involves an agent learning a policy -- a mapping from states to actions -- by receiving feedback in the form of a reward or a penalty. This feedback guides the agent in optimizing cumulative rewards over time. The key elements of RL include an agent, an environment, a set of states and actions, and a reward function. At each step, the agent observes the current state $s$ of the environment and selects an action $a$ from the action set $A$. Upon executing action $a$, the environment transitions to a new state $s'$, and the agent receives feedback in the form of a reward or a penalty. Throughout the learning process, the agent aims to maximize the cumulative reward by iteratively refining its strategy through trial-and-error interactions with the environment.

Classical RL methods require a complete model of the environment, which includes knowledge of all possible states, the set of actions available under each state, the transition probability matrix, and the expected reward values to estimate the total cumulative reward an agent can achieve. However, such comprehensive models are often unavailable, particularly for COPs \citep{wauters2013boosting}. In these situations, algorithms like Monte Carlo and Temporal Difference methods are employed \citep{sutton2018reinforcement}. Q-learning, a model-free RL algorithm based on temporal differences \citep{watkins1989learning}, addresses this challenge by associating each state-action pair $(s, a) \in S \times A $ with a score $Q(s,a) \in \mathbb{R}$, namely Q-value, which represents the expected reward from choosing action $a$ at state $s$:

\begin{equation} \label{eq: Q_value function}
    Q(s,a) := Q(s,a) + \alpha [r + \gamma \max_{a'} Q(s',a') - Q(s,a)] \;,
\end{equation}
\noindent

\noindent where $s$ is the current state, $a$ is the action taken in state $s$, $s'$ is the next state after performing action $a$, and $a'$ is a possible action in state $s'$. Furthermore, $\alpha$ ($0 < \alpha \leq 1$) controls the learning rate that determines the ratio of accepting newly learned information, $r$ is the reward received after performing action $a$, and $\gamma$ ($0 < \gamma \leq 1$) is the discount factor that determines the influence of the future reward $\max_{a'} Q(s',a')$. Note that $S$ is a finite set, and possible actions $A$ may generally depend on the state.

A crucial aspect of the Q-learning algorithm is balancing exploration and exploitation when selecting actions. One approach is to consistently choose the action with the highest $Q$-value, thereby enhancing the exploitation aspect of the algorithm. However, this can lead to other state-action pairs being insufficiently explored. Q-learning is proven to converge to the optimal $Q(s,a)$ if each state-action pair is visited a sufficient number of times \citep{watkins1989learning}. Therefore, it is essential to explore other state-action pairs, not just those with the highest Q-values. The $\epsilon$-greedy strategy \citep{sutton2018reinforcement} effectively balances exploration and exploitation by assigning a probability $\epsilon$ to select alternative actions. This strategy is expressed as follows:

\begin{equation} \label{eq: epsilon-greedy}
   a =
  \begin{cases}
    \argmax \limits_{a \in A} Q(s,a) & \text{with probability} \hspace{0.2cm} 1-\epsilon\\
    \text{any action selected uniformly and randomly in $A$}  & \text{with probability} \hspace{0.2cm} \epsilon
  \end{cases}
\end{equation}

A common practice is to gradually reduce the value of $\epsilon$ during the search process using a parameter known as $\epsilon$-decay, denoted by $\beta$. This reduction is designed to shift the focus from exploring new actions to increasingly exploiting the best actions as the search progresses.

\section{The proposed operator management framework} \label{sec: QIG}

\begin{figure}[H]
    \centering
    \includegraphics[scale = 0.8]{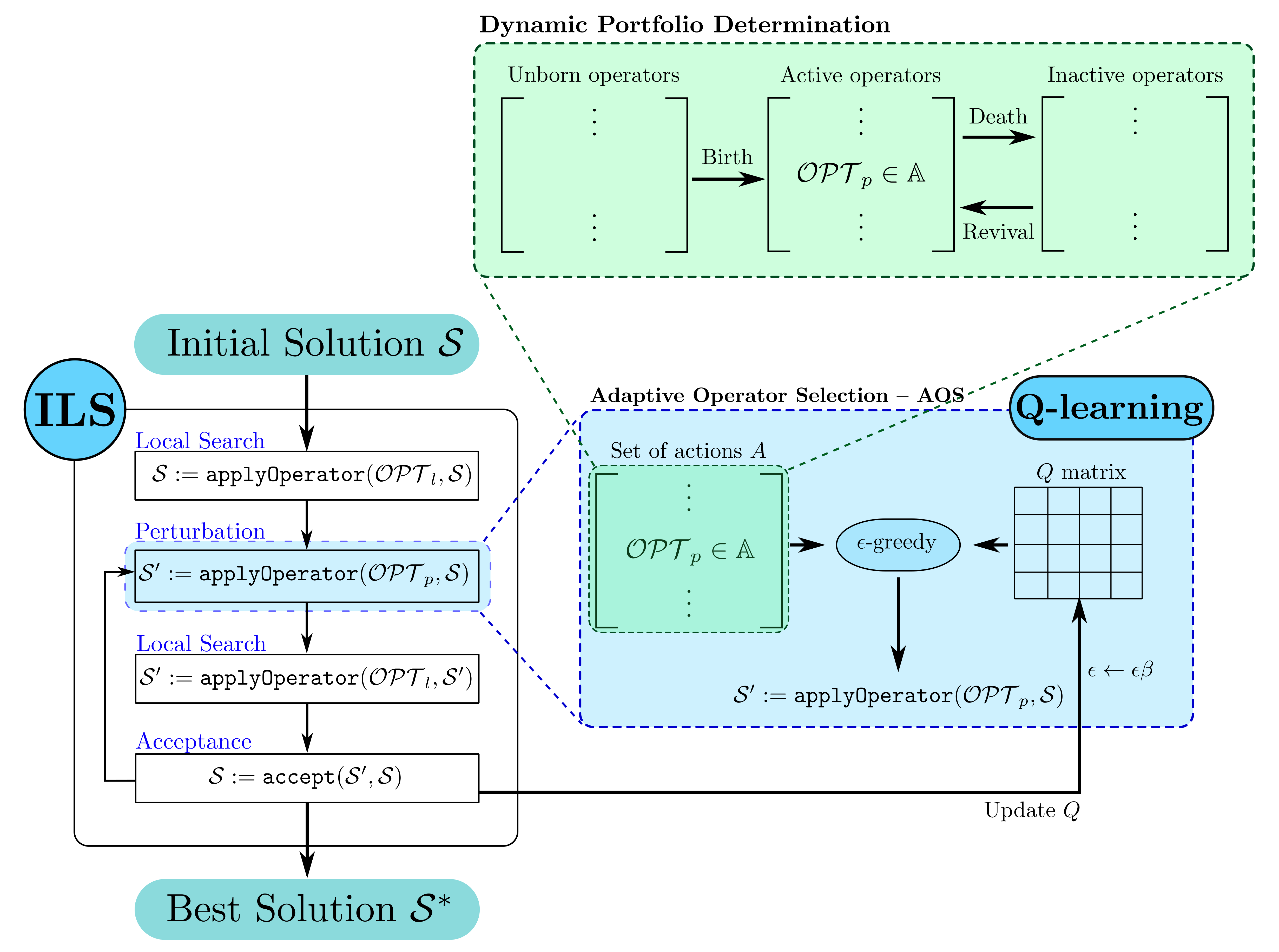}
    \caption{The procedure of the proposed operator management framework} 
    \label{fig:DQIG-scheme}    
\end{figure}

In this section, first, the general idea of the proposed framework is presented through Figure \ref{fig:DQIG-scheme} and Sections \ref{sec: portfolio determination} and \ref{sec: operator selection}. Then, the proposed framework is adapted to solve the PFSP in Section \ref{sec: DQIG for PFSP}.   

Figure \ref{fig:DQIG-scheme} illustrates the flowchart of the proposed framework applied to manage search operators within a general iterated local search (ILS) meta-heuristic. In this figure, we aim to manage a set of perturbation operators $\mathcal{OPT}_p$; however, we can easily adapt the framework to manage a set of local search operators $\mathcal{OPT}_l$.

The two phases of the proposed framework, namely dynamic portfolio determination and adaptive operator selection, shown in Figure \ref{fig:DQIG-scheme} are further explained in Sections \ref{sec: portfolio determination} and \ref{sec: operator selection}, respectively.

\subsection{Dynamic portfolio determination} \label{sec: portfolio determination}

The main goal of the dynamic portfolio determination in the proposed framework is to dynamically decide which operators to include in the portfolio for the subsequent AOS phase.

The pseudo-code of our dynamic portfolio determination mechanism is given as Algorithm \ref{alg: Tabu}. Inspired by the idea of tabu search \citep{glover1998tabu}, we design a portfolio determination mechanism that temporarily removes unsuccessful operators from the portfolio by moving them to a tabu list $\mathbb{T}$ for a period of further $\Theta$ episodes (i.e., tabu tenure), where each episode consists of a fixed number of iterations. This prevents unsuccessful operators from being selected by the AOS phase. After passing $\Theta$ episodes in the tabu list, the inactive operator returns to the portfolio based on the FIFO (First-In-First-Out) rule.  Consequently, as shown in Figure \ref{fig:DQIG-scheme}, an operator may have either of these three states: \textit{unborn}, \textit{active}, or \textit{inactive}. Unborn operators are those operators that are accessible (reserved) but not present in the portfolio. Active operators are the operators present in the portfolio from which an operator is selected by Q-learning in the AOS phase. Inactive operators are those unsuccessful operators identified based on their short-term performance in the most recent episode and temporarily removed from the portfolio. It is worth mentioning that removing an operator from the portfolio does not imply that the operator is inherently inefficient; rather, it reflects its inefficiency in the current stage. An inactive operator at this stage may prove valuable in the subsequent stages; therefore, inactive operators are only temporarily excluded. 

To be more specific, at the end of each episode, the portfolio is updated based on the short-term performance of the operators in the most recent episode. If an operator selected by AOS is unable to improve the current solution, it becomes inactive and temporarily leaves the portfolio (death transition) toward the tabu list. An inactive operator stays $\Theta$ episodes in the tabu list and then returns to the portfolio (revival transition) based on the FIFO (First-In-First-Out) rule. 

\vspace{0.1cm}
\SetAlgoCaptionSeparator{.}
\IncMargin{1em}

\begin{algorithm}[H]
\caption{Pseudo-code of the dynamic portfolio determination}
\label{alg: Tabu}

\SetKwFunction{FMain}{\texttt{determinePortfolio}}
\SetKwProg{Fn}{Function}{:}{}

\newcommand\mycommfont[1]{\footnotesize\ttfamily\textcolor{blue}{#1}}
\SetCommentSty{mycommfont}


\DontPrintSemicolon
\BlankLine
\KwIn{$A, a, \mathbb{T}, \theta_a, \Theta$}

\KwOut{$A, \mathbb{T}, \theta_a$ 
}

            
            


\BlankLine

\Fn{\FMain{$A, a, \mathbb{T}, \theta_a, \Theta$}}{
        \tcp*[l]{Death transition}
        \If{\upshape $r=0$}{
         $A := A \setminus a$ \tcp*[r]{Excluding unsuccessful action $a$ from list $A$}
        
         $\mathbb{T} := \mathbb{T} + \{a\}$ \tcp*[r]{Moving unsuccessful action $a$ into tabu list $\mathbb{T}$}

         $\theta_a := \Theta$ \tcp*[r]{Death of action $a$ for next $\Theta$ episodes}
        }
        
        \tcp*[l]{Revival transition}
         \For{$a \in \mathbb{T}$}{
         $\theta_a := \theta_a - 1$ 
            
             \If{\upshape $ \theta_a = 0$}{
             
              $A := A + \{a\}$ \tcp*[r]{Revival of $a$} 
            
             $\mathbb{T} := \mathbb{T} \setminus a$
             }}



    
    


        


    
    

    
    \Return $A, \mathbb{T}, \theta_a$
    
}

\end{algorithm}
\DecMargin{1em}
\vspace{0.1cm}

\subsection{Adaptive operator selection} \label{sec: operator selection}

This study proposes an AOS based on Q-learning, given as Algorithm \ref{alg: Q-learning}, where the action set $A$ corresponds to the set of active operators to be selected and applied at each episode. The state set is defined as $S = \{0, 1\}$, where $s \in S$  indicates whether the algorithm is trapped in a local optimum or not. $s = 0$ represents a situation where the search has been stuck in a local optimum, and $s = 1$ represents otherwise. 

As shown in Algorithm \ref{alg: Q-learning}, at the end of an episode, an operator is selected by Q-learning for the next episode from the updated portfolio using the $\epsilon$-greedy strategy. This selection is made based on both the operators' performance history and the search's current state. A selected operator is given one episode, equivalent to a number $E$ of iterations, to improve the solution. Then, the selected operator receives immediate feedback, i.e., a reward, from the environment. This feedback is a quantitative measure of the improvement achieved by applying the selected operator. Once the reward is assigned to the selected operator, the Q-value associated with the current state-action pair is updated. At this point, the search state is also updated. Finally, based on the value of the reward, the operator portfolio is updated. If the reward $r>0$, then the selected operator remains in the portfolio; otherwise, the operator leaves the portfolio and enters the tabu list. 

We define a reward function by considering the proportional improvement in the objective function, both locally and globally. The local reward is the amount of improvement in the current local optimum achieved through one full episode and is calculated in Equation (\ref{equ: local reward}). The global reward is the amount of improvement in the best solution found through one full episode and is calculated in Equation (\ref{equ: global reward}). Using a weight parameter for local/global reward, we then define a reward function as Equation (\ref{equ: total reward}), where $\eta$ represents the weights (importance) of a local improvement. 1-$\eta$ is, therefore, the weight of a global improvement.  

\begin{equation} \label{equ: local reward}
    r_{local} = \frac{\max ( \underline{C} - C, 0 )}{\underline{C}}
\end{equation}

\begin{equation} \label{equ: global reward}
    r_{global} = \frac{\max ( \underline{C}^* - C^*, 0 )}{\underline{C}^*}
\end{equation}

\begin{equation} \label{equ: total reward}
    r = \eta \cdot r_{local} + (1-\eta) \cdot r_{global}
\end{equation}

By updating the Q-values associated with state-action pairs through the Q-learning algorithm, the system learns which operators are most effective in various states, thus enabling an adaptive strategy that balances the exploration of new operators with the exploitation of known good operators. To balance exploration and exploitation in selecting the actions, we use the $\epsilon$-greedy strategy.

\vspace{0.1cm}
\SetAlgoCaptionSeparator{.}
\IncMargin{1em}
\def\HiLi{\leavevmode\rlap{\hbox to \hsize{\color{blue!30}\leaders\hrule height .6\baselineskip depth .5ex\hfill}}}

\begin{algorithm}[H]
\caption{Pseudo-code of the Q-learning-based AOS}
\label{alg: Q-learning}

\SetKwFunction{FMain}{\texttt{selectOperator}}
\SetKwProg{Fn}{Function}{:}{}

\newcommand\mycommfont[1]{\footnotesize\ttfamily\textcolor{blue}{#1}}
\SetCommentSty{mycommfont}


\DontPrintSemicolon
\BlankLine
\KwIn{$\underline{C}, \underline{C}^*, C, C^*, \epsilon,  \alpha, \beta, \gamma, \eta, A, s, a, \theta_a, \Theta, \mathbb{T}$}

\KwOut{$Q \in \mathbb{R}^{|S| \times |A|}, s', a', A, \mathbb{T}, \theta_a $ 
}


\BlankLine

\Fn{\FMain{$\underline{C}, \underline{C}^*, C, C^*, \epsilon, \alpha, \beta, \gamma, \eta, A, s, a, \theta_a, \Theta, \mathbb{T}$}}{

    \tcp*[l]{Calculate reward based on Equations (\ref{equ: local reward})--(\ref{equ: total reward})}
    $r := \texttt{calculateReward}(\underline{C}, \underline{C}^*, C, C^*, \eta)$

    \BlankLine

    \tcp*[l]{Update the state}
    \uIf{\upshape $C^* < \underline{C}^*$}{
        $s' := 1$
    }\Else{
        $s' := 0$
    }
    
    \BlankLine
    
    \tcp*[l]{Update $Q$ table}
    $Q(s,a) := Q(s,a) + \alpha \left[r + \gamma \max \limits_{a' \in A}Q(s',a') - Q(s,a)\right]$

    \BlankLine

    \tcp*[l]{Update portfolio}
     $A, \mathbb{T}, \theta_a:= \texttt{determinePortfolio}(A, a,  \mathbb{T}, \theta_a, \Theta)$
    
        


    \BlankLine
    
    \tcp*[l]{Select an action}
    \uIf{\upshape $\texttt{rand()} \geq \epsilon$}{
        $a' := \argmax \limits_{a'' \in A} Q(s',a'')$
    }\Else{
        $a' := \texttt{randomChoice}(A)$
    }
    
    \tcp*[l]{Apply epsilon-decay}
    $\epsilon := \epsilon \times \beta$

    \BlankLine
    
    \Return $Q, s', a', A, \mathbb{T}, \theta_a$
    
}

\end{algorithm}
\DecMargin{1em}
\vspace{0.1cm}

\subsection{Proposed operator management framework for PFSP} \label{sec: DQIG for PFSP}

In this section, the proposed operator management framework of Figure \ref{fig:DQIG-scheme} is applied to manage a large set of perturbation operators within the original IG algorithm (Algorithm \ref{alg: IG}) to solve the PFSP. Hereafter, we call the proposed framework DQIG (Dynamic Q-learning for Iterated Greedy). The pseudo-code of DQIG is provided as Algorithm \ref{alg: proposed DQIG}, wherein the differences between the proposed DQIG and the original IG are highlighted.  

As explained in Section \ref{sec: original IG}, the perturbation operator of the original IG is composed of two steps: a destruction step where $d$ jobs are randomly removed from the sequence and a construction phase where the removed jobs are inserted back into the partial sequence. Varying the value of $d$ leads to distinct destruction operators with different levels of destruction. While smaller values of $d$ result in minor disruptions, favoring local search, larger values of $d$ induce significant changes, promoting broader exploration of the solution space. In the construction phase, different construction strategies exist to insert back the removed jobs into the partial sequences, among which the most comment strategies include \citep{fernandez2017new} i) \textit{best-insertion}, ii) \textit{random-insertion}, iii) \textit{semi-random insertion}, and iv) \textit{probabilistic-insertion}. These construction strategies mainly differ in the level of randomness for reinserting the jobs. The \textit{best-insertion} strategy is the greediest strategy that inserts the jobs into their best position, yielding the minimum makespan for the partial sequence, focusing on intensification by exploiting known information. The \textit{probabilistic-insertion} strategy inserts the jobs into positions that are selected probabilistically, based on the value of the makespan, privileging the best reinsertion positions without fully sacrificing randomness. The \textit{semi-random insertion} strategy inserts part of the jobs into their best positions and the remaining part into completely random positions, making a compromise between full exploration and full exploitation. The \textit{random-insertion} strategy inserts jobs into completely random positions, introducing high variability into the solution construction step and promoting only full exploration. 

\vspace{0.3cm}
\SetAlgoCaptionSeparator{.}
\IncMargin{1em}
\def\HiLi{\leavevmode\rlap{\hbox to \hsize{\color{blue!30}\leaders\hrule height .6\baselineskip depth .5ex\hfill}}}

\begin{algorithm}[H]
\caption{Pseudo-code of the proposed DQIG algorithm}
\label{alg: proposed DQIG}

\SetKwFunction{FMain}{DQIG}
\SetKwProg{Fn}{Function}{:}{}

\newcommand\mycommfont[1]{\footnotesize\ttfamily\textcolor{blue}{#1}}
\SetCommentSty{mycommfont}


\DontPrintSemicolon

\HiLi{\KwIn{$\epsilon$, $\alpha$, $\beta$, $\gamma$, $E$, $\eta$, $\Theta$, $A$}}


\KwOut{$\Pi^* \in \mathfrak{S}(N)$ 
}


\Fn{\FMain{$\epsilon, \alpha, \beta, \gamma, E, \eta, \Theta, A$}}{

    
    $\Pi := \texttt{generateInitialSolution}()$ 
    
    $\Pi := \texttt{applyLocalSearch}(\Pi)$ 
    
    $\Pi^* := \Pi$ \tcp*[r]{Remember the best solution found}
    
    \HiLi $Q := [0]$ \tcp*[r]{Initialize $Q$, a zero-filled $|S| \times |A|$ table}
    
    \HiLi $s := 0$ \tcp*[r]{Initial state is fixed to 0}
    
    \HiLi $a := \texttt{randomChoice}(A)$ \tcp*[r]{Action for the first iteration is drawn randomly from $A$}

    \HiLi $\mathbb{T} := \{\}$ \tcp*[r]{All operators are active at the beginning of the search, and the tabu list $\mathbb{T}$ is empty}
    
    \HiLi $\theta_{a \in A} := 0$
    
    \While{\upshape \texttt{terminationCriterion}()}{

        
        \HiLi \tcp*[l]{Start of an episode}
        \HiLi $\underline{C} := C_{max}(\Pi)$ \tcp*[r]{Record $C_{max}$ of current local optimum prior to an episode}
        \HiLi $\underline{C}^* := C_{max}(\Pi^*)$ \tcp*[r]{Record $C_{max}$ of the best solution found prior to an episode}
        \HiLi $C := \underline{C}$ \tcp*[r]{Record $C_{max}$ of the best local optimum during an episode}
        \HiLi $C^* := \underline{C}^*$ \tcp*[r]{Record $C_{max}$ of the best solution found during an episode}
        \For{$e = 1: E$}{
            
            
            \tcp*[l]{\textbf{Perturbation}}
            $D := \texttt{destructSolution}(\Pi,a)$ 
            
            $\Pi_{\setminus D} := \Pi \setminus D$ 
            
            $\Pi_{\setminus D} := \texttt{applyLocalSearch}(\Pi_{\setminus D})$
            
            $\Pi' := \texttt{constructSolution}(\Pi_{\setminus D}, D, a)$ 
    
            \BlankLine
    
            \tcp*[l]{\textbf{Local search}}
            $\Pi' := \texttt{applyLocalSearch}(\Pi')$
    
            \BlankLine
    
            \tcp*[l]{\textbf{Acceptance}}
            $\Pi := \texttt{acceptSolution}(\Pi', \Pi)$

            {

                \HiLi $C := \min(C, C_{max}(\Pi))$ 
            }
            
            $\Pi^* := \texttt{updateBestSolution}(\Pi', \Pi^*)$ 

            \HiLi $C^* := C_{max}(\Pi^*)$

            

        }
            
        
        \tcp*[l]{Change state $s$, update $Q$, update $A$ and $\mathbb{T}$, and select $a$ (action) for next episode}
        \HiLi $Q, s, a := \texttt{selectOperator}(\underline{C}, \underline{C}^*, C, C^*, \epsilon, \alpha, \beta, \gamma, \eta, A, s, a, \theta_a, \Theta, \mathbb{T})$


            
            

         }

        
    }
    
    
    \Return $\Pi^*$

\end{algorithm}
\DecMargin{1em}
\vspace{0.3cm}

Depending on the size of instances to determine the value of $d$ and the variety of construction strategies, one can generate a large number of perturbation operators (i.e., $(d, \texttt{construction strategy})$ tuple). To efficiently manage a large number of operators, the proposed DQIG framework deploys the tabu idea and Q-learning to update the portfolio and select the appropriate $(d, \texttt{construction strategy})$ tuples, respectively, at each stage. Our framework has two prominent advantages compared to static portfolio determination: removing the need to pre-determine a fixed portfolio of operators (i.e., zero pre-processing effort) and reducing the chance of employing inefficient operators at every stage of the search while keeping all operators accessible throughout the search process whenever needed.

\subsection{Complexity of the proposed DQIG framework} \label{sec: complexity}

Embedding Q-learning and the idea of tabu within the original IG results in a computational overhead. In the following, we quantify this overhead and show that it is independent of the instance size. The worst-case complexity of the proposed DQIG framework is therefore analyzed according to the functions used in Algorithm \ref{alg: proposed DQIG} and the choices made for each function as follows:

\begin{itemize}
    \itemsep0em
    \item \texttt{generateInitialSolution}: In this work, the Nawaz, Enscore, and Ham (NEH) heuristic \citep{nawaz1983heuristic} based on Taillard's acceleration method \citep{taillard1990some} is used to generate initial solutions. The complexity of this function is $O(n^2m)$ as can be derived from Algorithm \ref{alg: NEH} in \ref{apendix: NEH}. 
    \item \texttt{applyLocalSearch}: In this work, the insertion neighborhood is used as the local search. The local search is also applied to the partial sequence obtained after the destruction step \citep{dubois2017iterated}. The complexities of insertion neighborhood local search for complete and partial sequences are $O(n^2m)$ and $O((n-d)^2m)$, respectively \citep{ruiz2007simple} as can be derived from Algorithm \ref{alg: insertLocalSearch} in \ref{apendix: NEH}. 
    \item \texttt{terminationCriterion}: Different termination criteria can be used, such as the total execution time, the number of iterations, or the number of non-improving iterations. In any of such cases, this can be done in $O(1)$.
    \item \texttt{destructSolution}: This function randomly extracts $d$ jobs from the current solution, which results in a remaining partial sequence. This operation can be done in $O(d)$.
    \item \texttt{constructSolution}: In this work, four operators for constructing the solution are used: random-insertion, semi-random insertion, probabilistic-insertion, and best-insertion. The worst-case complexity among all these operators is $O(dnm)$ \citep{ruiz2007simple}.
    \item \texttt{acceptSolution}: In this work, Metropolis acceptance strategy \citep{metropolis1953equation} is used to accept/reject a solution. If the solution improves the objective function, it is always accepted. If it is a non-improving solution, it is accepted with a probability of $\texttt{exp}({\frac{C_{max}(\Pi)-C_{max}(\Pi')}{T_p}})$ where $T_p$ is a cooling temperature parameter. $T_p$ is calculated as Equation (\ref{equ: temperature}), wherein $\tau$ is the temperature scale \citep{ruiz2007simple}. The complexity of this function is $O(1)$.
    \begin{equation} \label{equ: temperature}
    T_p = \tau \cdot \frac{\sum_{i=1}^n\sum_{j=1}^m p_{ij}}{n \cdot m \cdot 10}
    \end{equation} 
    \item \texttt{updateBestSolution}: The complexity of this function is $O(1)$.
    \item \texttt{selectOperator}: This function encompasses the following operations:
    \begin{itemize}
        \item Q-learning selects an operator where the only required operation is to find a maximum $Q$-value from a list of size $|A|$. The complexity of this operation is $O(|A|)$.
        \item \texttt{determinePortfolio}: The complexity of this function is $O(|\mathbb{T}|)$.
    \end{itemize}
    
    \item A tie-breaking mechanism proposed by \cite{fernandez2014insertion} has been used in generateInitialSolution, applyLocalSearch, and constructSolution functions. This tie-breaking mechanism prioritizes jobs with the lowest sum of the idle times in the sequence. Using this tie-breaking mechanism does not affect the complexity of the corresponding functions.
\end{itemize}

To assess the computational overhead of the proposed DQIG framework compared to the original IG, we now determine the worst-case complexities of both the original IG of Algorithm \ref{alg: IG} and the proposed DQIG of Algorithm \ref{alg: proposed DQIG}. For this purpose, the same termination criterion, the maximum number of iterations, is considered for both the original IG and the proposed DQIG. Let $I_{IG}$ and $I_{DQIG}$ denote the number of iterations performed by the \texttt{while} loops of IG (line 5 of Algorithm \ref{alg: IG}) and DQIG (line 11 of Algorithm \ref{alg: proposed DQIG}), respectively. As we use the concept of episodes in DQIG, the number of iterations in the \texttt{while} loop is multiplied by the number of iterations in the \texttt{for} loop (line 17 of Algorithm \ref{alg: proposed DQIG}). As a result, ($I_{IG}=I_{DQIG}\times E$).

The worst-case complexities of the original IG and the proposed DQIG are given as Equations \ref{equ: complexIG} and \ref{equ: complexDQIG}, respectively.

\begin{equation} \label{equ: complexIG}
    O(2n^2m + I_{IG} (d + (n-d)^2m + dnm + n^2m))
\end{equation}

\begin{equation} \label{equ: complexDQIG}
    O(2n^2m + I_{DQIG} (E( d + (n-d)^2m + dnm + n^2m) + |A| + |\mathbb{T}|))
\end{equation}

Comparing Equation \ref{equ: complexDQIG} to \ref{equ: complexIG}, the complexity overhead of DQIG is:

\begin{equation}
    O(I_{DQIG} \times (|A|+|\mathbb{T}|))
\end{equation}

Accordingly, the complexity overhead of DQIG is a constant that does not grow with the size of the problem instance. This overhead only imposes $O(|A|+|\mathbb{T}|) = O(1)$ extra computations per iteration. In Section \ref{sec: numerical and statistical results}, we show that DQIG finds better solutions in practice than the original IG and the state-of-the-art algorithms under the same computational time. This is attributed to its faster convergence toward better solutions, which is discussed in Section \ref{sec: numerical and statistical results}. 

\section{Experimental design} \label{sec: Experimental design}
In this section, we design a full set of experiments to investigate the performance of the proposed DQIG framework. 

The computational experiments are conducted on two widely recognized PFSP datasets: the \textit{Taillard} dataset \citep{taillard1993benchmarks}, which comprises 120 instances divided into 12 sets, ranging from 20 jobs with 5 machines to 500 jobs with 20 machines, and the \textit{VRF-hard-large} dataset \citep{vallada2015new}, which includes 240 instances divided into 24 sets, ranging from 100 jobs with 20 machines to 800 jobs with 60 machines. To provide a fair comparison between DQIG and all the benchmark algorithms, the stopping criterion for all algorithms is set to a limited computational time. In the PFSP literature, the computational time for each instance is calculated based on the instance size as $T=\frac{nm}{2}t$ milliseconds (ms) where $t$ is a timescale ($t\in \{60, 90, 120\}$) \citep{ruiz2007simple}. It is important to note that all algorithms were implemented and tested under identical conditions. They were coded in Python 3.7 and executed on the same computer environment. All experiments were conducted on a server equipped with eight Intel XEON processors running at 2.3 GHz with 5GB of RAM. Only a single processor was used for the experiments, and no parallel programming techniques were employed. The source codes of the framework have been made available online\footnote{Peer reviewing anonymous link: \url{https://osf.io/c8ut9/?view_only=26ac9117ff1e4108ba04fa48577ca7ec}} to ensure reproducibility of the results.

\subsection{Experimental setting} \label{sec: experimental setting}
To analyze the performance of DQIG, first, the contribution of both the dynamic portfolio determination and the Q-learning-based adaptive operator selection mechanisms to the performance of DQIG is analyzed. Then, the competitiveness of DQIG with respect to the state-of-the-art algorithms is benchmarked. The details of the benchmark algorithms are provided as follows:

\begin{itemize}
    \itemsep0em 
    \item The contribution of the dynamic portfolio determination and the Q-learning-based AOS mechanisms (see Section \ref{sec: phase 1} for the numerical results) -- This is analyzed by comparing DQIG to its variants given as follows:  
    \begin{itemize}
        \item SQIG: A variant of DQIG where the only difference is that the portfolio determination is static \citep{karimi2023learning}, meaning that all perturbation operators are always active. This is to analyze the contribution of our dynamic portfolio determination mechanism;
        \item RIG: A variant of DQIG where the only difference is that the operator selection in AOS is random. This is to analyze the contribution of Q-learning in AOS.
        \item ScIG: A variant of DQIG where the portfolio determination is static, and the AOS mechanism is based on the classical score-based mechanism of the adaptive large neighborhood search (ALNS) \citep{aksen2014adaptive}. This is to analyze the contribution of the whole operator management mechanism.  
    \end{itemize}
    
    \item The competitiveness of DQIG (see Section \ref{sec: phase 2} for the numerical results) -- The competitiveness of the proposed DQIG framework is analyzed by comparing DQIG to five state-of-the-art IG algorithms from the literature for PFSP:
    \begin{itemize}
 
    \item IG$_{RS}$: The original IG algorithm employing a single perturbation operator ($d=4$, best insertion) \citep{ruiz2007simple};
    \item IG$_{DPS}$: An IG algorithm employing a single perturbation operator ($d=2$, best insertion) whose main idea is to apply an extra local search to the partial solutions after the destruction step \citep{dubois2017iterated};
    \item IG$_{KTPG}$: An IG algorithm employing a single perturbation operator ($d=4$, best insertion) whose main idea is to apply a variable-blocking perturbation mechanism, wherein a block of jobs of size $d = 2$ are simultaneously removed and inserted in the best position with the minimum makespan \citep{kizilay2019variable};  
    \item IG$_{FF}$: An IG algorithm employing a single perturbation operator ($d=2$, best insertion) whose main idea is to combine efficient features of previous IG algorithms from the literature to design a best-of-breed algorithm. In addition, a new tie-breaking mechanism called FF has been used \citep{fernandez2019best};
    \item IG$_{PS}$: An IG algorithm employing a single perturbation operator ($d=1$, best insertion) whose main idea is to automatically design flexible algorithms by selecting high-performing configurations for solving different problem instances \cite{pagnozzi2019automatic}.
    \end{itemize}
\end{itemize}

\subsection{Performance metrics}

To account for the stochastic nature of the algorithms, each algorithm was executed 30 independent times per instance. Two performance metrics are used to compare the performance of the algorithms \citep{fernandez2017new}: the Average Relative Percentage Deviation (ARPD) to measure the quality of the solutions and the average computational time (in seconds) to measure the required computational effort. ARPD measures the deviation from the proven optimum (if available) or the best-known solution in the literature \citep{kurdi2020memetic,vallada2015new}. The Relative Percentage Deviation (RPD) of algorithm $j$ over instance $i$, $RPD_{i,j}$, and the average relative performance deviation (ARPD) of algorithm $j$ are calculated as Equations (\ref{equ: RPD}) and (\ref{equ: ARPD}), respectively, where $C_{max,i}^*$ is the makespan of the optimal solution (or best-known solution) for instance $i$ belonging to the instance set $I$.

\begin{equation} \label{equ: RPD}
   RPD_{i,j} = \frac{C_{max,i,j}-C_{max,i}^*}{C_{max,i}^*} \times 100 \hspace{0.5cm} \forall i,j
\end{equation}

\begin{equation} \label{equ: ARPD}
   ARPD_{j} =  \frac{\sum_{i \in I}{RPD_{i,j}}}{|I|} \hspace{0.5cm} \forall j
\end{equation}

Furthermore, we use statistical tests to determine whether observed differences in algorithm performances are statistically significant. Given that the normality assumption (of the difference between the means of two algorithms) is not satisfied, we use the Wilcoxon signed-rank test \citep{wilcoxon1992individual} with a 95\% confidence level. The Wilcoxon signed-rank test is a non-parametric statistical test used to assess whether there is a significant difference between paired observations (i.e., algorithms).

\subsection{Parameters tuning} \label{sec: parameter tuning}

In this work, the response surface methodology (RSM) and the $3k$ factorial Box–Behnken design of experiment \citep{wang2009hybrid,zhalechian2018hub} are used to determine the best combinations of parameters for DQIG. In $3k$ factorial Box–Behnken method, each factor is tested at three levels; (min, mean, max) levels transformed into coded levels (-1, 0, 1). This transformation facilitates the analysis of the experimental design. To transform the levels of a factor into these standardized levels, Equation \ref{equ: coded variables} is used, where $X$ is the coded value of the factor, $x$ is its uncoded value, and $x_{max}$ and $x_{min}$ are the maximum and minimum levels of the factor.

\begin{equation} \label{equ: coded variables}
   X = \frac{x- \frac{x_{max}+x_{min}}{2}}{\frac{x_{max}-x_{min}}{2}} 
\end{equation}

DQIG has seven significant parameters (factors), shown in Table \ref{tab: parameter tuning}. This results in a total number of 62 experiments \citep{box1960some}. The experiments are conducted on a set of randomly selected instances from benchmark datasets (i.e., 12 instances from \textit{Taillard} and 24 instances from \textit{VRF-hard-large}) with a stopping criterion of $T= \frac{nm}{2}60$ ms. 
Table \ref{tab: parameter tuning} shows the best combination of the obtained parameters.

{
\fontsize{6}{7}\selectfont
\setlength{\tabcolsep}{0.5pt}

\begin{tabularx}{1.8\textwidth}{
                       >{\hsize=0.14\linewidth}X
                       >{\hsize=0.06\linewidth}X
                       >{\hsize=0.06\linewidth}X 
                       >{\hsize=0.06\linewidth}X 
                       >{\hsize=0.06\linewidth}X
                       >{\hsize=0.06\linewidth}X}

  \caption{The tuned values of DQIG's parameters}
  \label{tab: parameter tuning} \\
  
  \toprule
 Parameter & Notation& Levels & & & Tuned value \\
 \cmidrule(r){3-5}
  	& &	-1 & 0 & 1 &  \\
  \midrule
  \endfirsthead
  \multicolumn{5}{c}{\tablename~\thetable~ (continued)} \\
  \toprule
 Parameter & Notation & Levels & & & Tuned value \\
 \cmidrule(r){3-5}
  	& &	-1 & 0 & 1 &  \\
 
  \midrule
  \endhead
  \midrule
  \multicolumn{5}{r@{}}{\footnotesize To be continued ...}
  \endfoot
  \bottomrule
  \endlastfoot

Epsilon-greedy & $\epsilon$ & 0.7 & 0.85 & 1 & 0.8\\
Epsilon-decay & $\beta$ & 0.990 & 0.995  & 1 & 0.996 \\
Learning rate & $\alpha$ & 0.5 & 0.75 & 1 & 0.6\\
Discount factor & $\gamma$ & 0.7 & 0.85 & 1 & 0.8 \\
Size of episode & $E$ & 1 & 4 & 7 & 6\\
Local/global improvement weight & $\eta$ & 0.2 & 0.5 & 0.8 & 0.3 \\
Tabu tenure size & $\Theta$ & 2 & 4 & 6 & 4 \\

\end{tabularx}}

\section{Numerical and statistical results} \label{sec: numerical and statistical results}

\subsection{Comparison of DQIG with its variants} \label{sec: phase 1}

In this section, we compare the performance of DQIG with its variants (i.e., static portfolio, random operator selection, and score-based operator selection) under the same computational time. The results of these comparisons on Taillard dataset are provided in Table \ref{tab: QIG vs. Benchmarks scale 60 Taillard}. In this table, column ``Average'' reports the ARPD (\%) across 300 solutions (i.e., 10 instances of each size and 30 independent runs for each instance), and column ``Best'' shows the ARPD (\%) across the 10 best solutions (i.e., the best solution among 30 runs). The values in bold show the statistical significance of DQIG against each of the other algorithms with 95\% confidence interval.

According to Table \ref{tab: QIG vs. Benchmarks scale 60 Taillard}, DQIG consistently outperforms SQIG, indicating that dynamically (versus statically) adjusting the operator portfolio enhances the solution quality. The results highlight how temporarily inactivating unsuccessful operators allows the DQIG to focus on more effective operators, leading to improved search performance. Comparing DQIG to RIG, the results demonstrate that DQIG outperforms RIG, highlighting the effectiveness of the Q-learning-based AOS. Unlike random selection, Q-learning leverages the history of performance, the future gain, and the state of the search to prioritize operators with higher probabilities of success, thereby guiding the search process more efficiently and enhancing overall search performance. Comparing DQIG to ScIG, the results demonstrate that DQIG outperforms ScIG, highlighting the efficiency of the proposed operator management mechanism. Unlike score-based mechanisms that rely solely on the cumulative past performance of a static portfolio of operators, the proposed operator management framework adjusts the portfolio and the selection of operators to the state of the search based on the performance history and the future gains of operators. This balance between exploiting past success and anticipating future potential leads to significant improvements in solution quality over traditional score-based mechanisms. Comparing the results of DQIG versus its variants across different time scales $t=60, 90, 120$, it is evident that the DQIG exhibits a faster convergence to better solutions from the beginning, and this superior performance is consistently maintained throughout the search. The outperformance of DQIG against its variants is statistically significant for most instance sets. For certain instance sets, such as ``tai\_20\_10'', ``tai\_20\_20'', and ``tai\_50\_5'', while DQIG achieves better results compared to other algorithms, the observed differences are not statistically significant. This is primarily because these instance sets are relatively simple PFSP problems where all algorithms are able to find solutions with small optimality gaps.

{
\fontsize{6}{7}\selectfont
\setlength{\tabcolsep}{0.5pt}

\begin{tabularx}{2.2\textwidth}{
                       >{\hsize=0.03\linewidth}X
                       >{\hsize=0.03\linewidth}X 
                       >{\hsize=0.045\linewidth}X 
                       >{\hsize=0.045\linewidth}X
                       >{\hsize=0.045\linewidth}X 
                       >{\hsize=0.045\linewidth}X
                       >{\hsize=0.045\linewidth}X 
                       >{\hsize=0.045\linewidth}X
                       >{\hsize=0.045\linewidth}X 
                       >{\hsize=0.045\linewidth}X}

  \caption{Performance comparison of DQIG against its variants on \textit{Taillard} dataset for $t=60, 90, 120$}

  \label{tab: QIG vs. Benchmarks scale 60 Taillard} \\
  
  \toprule
  \multicolumn{2}{l}{Instance}	&	\multicolumn{8}{l}{Algorithms ($t=60$)}	\\
  \hline
  && \multicolumn{2}{l}{\text{ScIG}} &	\multicolumn{2}{l}{SQIG}	& \multicolumn{2}{l}{$\text{RIG}$}		&	\multicolumn{2}{l}{DQIG}\\  
  \hline
  $n$ & $m$&	Average	&	Best	&	Average	&	Best	&	Average	&	Best	&	Average	&	Best\\

  \midrule
  \endfirsthead
  \multicolumn{10}{c}{\tablename~\thetable~ (continued)} \\
  \toprule
  \multicolumn{2}{l}{Instance}	&	\multicolumn{8}{l}{Algorithms}	\\
  \hline
  && \multicolumn{2}{l}{\text{ScIG}} &	\multicolumn{2}{l}{SQIG}	& \multicolumn{2}{l}{$\text{RIG}$}		&	\multicolumn{2}{l}{DQIG}\\  
  \hline
  $n$ & $m$&	Average	&	Best	&	Average	&	Best	&	Average	&	Best	&	Average	&	Best\\

  \midrule
  \endhead
  \midrule
  \multicolumn{10}{r@{}}{\footnotesize To be continued ...}
  \endfoot
  \bottomrule
  \endlastfoot

20	&	5	&	0.032	&	0	&	0.038	&	0	&	0.041	&	0.041	&	\textbf{0.020}	&	0	\\
20	&	10	&	0	&	0	&	0.003	&	0	&	0.007	&	0	&	0	&	0	\\
20	&	20	&	\textbf{0.003}	&	0	&	0.006	&	0	&	\textbf{0.003}	&	0	&	\textbf{0.002}	&	0	\\
50	&	5	&	0.001	&	0	&	0.001	&	0	&	0.001	&	0	&	0.001	&	0	\\
50	&	10	&	0.550	&	0.443	&	\textbf{0.511}	&	0.286	&	\textbf{0.513}	&	0.398	&	\textbf{0.494}	&	0.307	\\
50	&	20	&	0.693	&	0.533	&	\textbf{0.614}	&	0.264	&	0.646	&	0.509	&	\textbf{0.611}	&	0.278	\\
100	&	5	&	0.009	&	0.008	&	0.011	&	0.008	&	\textbf{0.008}	&	0.008	&	\textbf{0.008}	&	0.008	\\
100	&	10	&	0.108	&	0.030	&	\textbf{0.097	}&	0.021	&	\textbf{0.092}	&	0.039	&	\textbf{0.091}	&	0.021	\\
100	&	20	&	0.950	&	0.783	&	\textbf{0.929}	&	0.599	&	0.938	&	0.707	&	\textbf{0.897}	&	0.569	\\
200	&	10	&	0.045	&	0.032	&	0.052	&	0.037	&	0.049	&	0.033	&	0.049	&	0.030	\\
200	&	20	&	0.883	&	0.700	&	0.908	&	0.595	&	\textbf{0.869}	&	0.696	&	\textbf{0.839}	&	0.576	\\
500	&	20	&	0.406	&	0.332	&	0.366	&	0.235	&	0.387	&	0.317	&	\textbf{0.338}	&	0.223	\\

\\

\multicolumn{2}{l}{Average} & 0.307	&	0.238	&	0.295	&	0.170	&	0.296	&	0.229	&	\textbf{0.279}	&	0.168	\\

\hline
\\

\multicolumn{2}{l}{Instance}	&	\multicolumn{8}{l}{Algorithms ($t=90$)}	\\
  \hline
  && \multicolumn{2}{l}{\text{ScIG}} &	\multicolumn{2}{l}{SQIG}	& \multicolumn{2}{l}{$\text{RIG}$}		&	\multicolumn{2}{l}{DQIG}\\  
  \hline
  $n$ & $m$&	Average	&	Best	&	Average	&	Best	&	Average	&	Best	&	Average	&	Best\\
\hline

\\

20	&	5	&	\textbf{0.024}	&	0	&	0.035	&	0	&	0.041	&	0.041	&	\textbf{0.020}	&	0	\\
20	&	10	&	0	&	0	&	0	&	0	&	0.007	&	0	&	0	&	0	\\
20	&	20	&	0.003	&	0	&	0.002	&	0	&	0	&	0	&	0	&	0	\\
50	&	5	&	0.000	&	0	&	0.000	&	0	&	0	&	0	&	0	&	0	\\
50	&	10	&	0.512	&	0.356	&	\textbf{0.472}	&	0.283	&	0.484	&	0.377	&	\textbf{0.456}	&	0.293	\\
50	&	20	&	0.633	&	0.465	&	\textbf{0.543}	&	0.256	&	0.615	&	0.479	&	\textbf{0.540}	&	0.240	\\
100	&	5	&	0.008	&	0.008	&	0.009	&	0.008	&	0.008	&	0.008	&	0.008	&	0.008	\\
100	&	10	&	0.068	&	0.025	&	\textbf{0.066}	&	0.021	&	\textbf{0.068}	&	0.025	&	\textbf{0.060}	&	0.021	\\
100	&	20	&	0.855	&	0.680	&	\textbf{0.833}	&	0.554	&	0.866	&	0.674	&	\textbf{0.812}	&	0.476	\\
200	&	10	&	0.042	&	0.031	&	0.047	&	0.037	&	0.042	&	0.033	&	\textbf{0.042}	&	0.036	\\
200	&	20	&	0.826	&	0.665	&	0.831	&	0.531	&	0.794	&	0.677	&	\textbf{0.767}	&	0.528	\\
500	&	20	&	0.387	&	0.323	&	0.345	&	0.231	&	0.367	&	0.299	&	\textbf{0.316}	&	0.211	\\

\\

\multicolumn{2}{l}{Average} &   	0.280	&	0.213	&	0.265	&	0.160	&	0.274	&	0.218	&	\textbf{0.252}	&	0.151	\\

\hline
\\

\multicolumn{2}{l}{Instance}	&	\multicolumn{8}{l}{Algorithms ($t=120$)}	\\
  \hline
  && \multicolumn{2}{l}{\text{ScIG}} &	\multicolumn{2}{l}{SQIG}	& \multicolumn{2}{l}{$\text{RIG}$}		&	\multicolumn{2}{l}{DQIG}\\  
  \hline
  $n$ & $m$&	Average	&	Best	&	Average	&	Best	&	Average	&	Best	&	Average	&	Best\\
\hline

\\

20	&	5	&	\textbf{0.024}	&	0	&	0.032	&	0	&	0.041	&	0.041	&	\textbf{0.020}	&	0	\\
20	&	10	&	0	&	0	&	0	&	0	&	0.007	&	0	&	0	&	0	\\
20	&	20	&	0	&	0	&	0.002	&	0	&	0	&	0	&	0	&	0	\\
50	&	5	&	0	&	0	&	0	&	0	&	0	&	0	&	0	&	0	\\
50	&	10	&	0.490	&	0.352	&	\textbf{0.438}	&	0.279	&	0.461	&	0.368	&	\textbf{0.431}	&	0.293	\\
50	&	20	&	0.600	&	0.444	&	\textbf{0.503}	&	0.234	&	0.571	&	0.418	&	\textbf{0.483}	&	0.202	\\
100	&	5	&	0.008	&	0.008	&	0.008	&	0.008	&	0.008	&	0.008	&	0.008	&	0.008	\\
100	&	10	&	0.058	&	0.025	&	0.051	&	0.019	&	0.060	&	0.025	&	\textbf{0.043}	&	0.019	\\
100	&	20	&	0.819	&	0.634	&	\textbf{0.782}	&	0.518	&	0.824	&	0.637	&	\textbf{0.759}	&	0.436	\\
200	&	10	&	0.040	&	0.031	&	0.043	&	0.036	&	0.039	&	0.033	&	\textbf{0.038}	&	0.036	\\
200	&	20	&	0.783	&	0.639	&	0.779	&	0.515	&	0.764	&	0.646	&	\textbf{0.714}	&	0.468	\\
500	&	20	&	0.366	&	0.305	&	0.330	&	0.231	&	0.349	&	0.281	&	\textbf{0.301}	&	0.204	\\

\\

\multicolumn{2}{l}{Average} & 	0.266	&	0.203	&	0.247	&	0.153	&	0.260	&	0.205	&	\textbf{0.233}	&	0.139	\\

\end{tabularx}}

\begin{figure}[H]
    \centering
    \includegraphics[scale = 0.35]{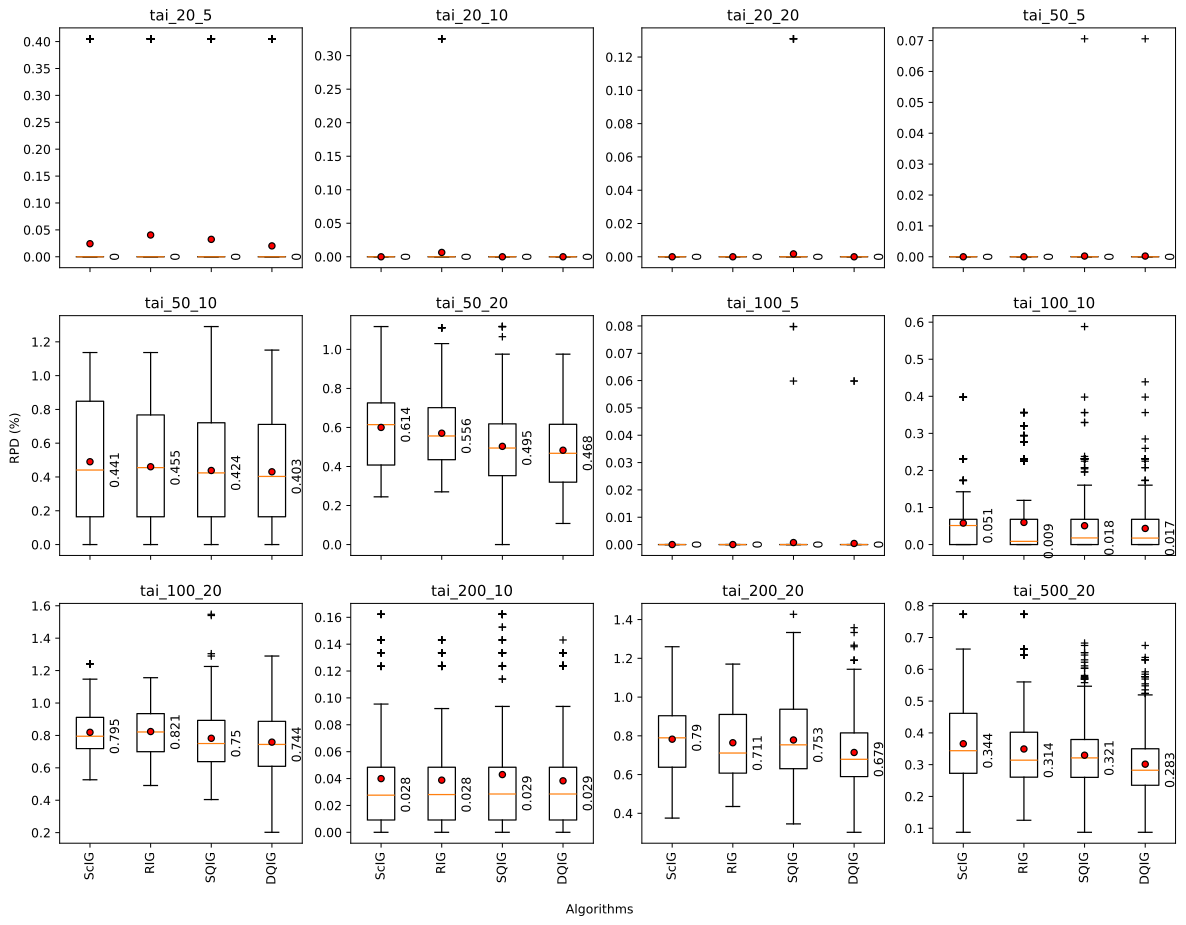}
    \caption{Boxplot of DQIG and its variants based on RPD (\%) for scale $t=120$ for each instance set of \textit{Taillard} dataset. The mean of each algorithm is shown using red bullets, and the median value is given on the right side of each boxplot.} 
    \label{fig:boxplot_time_Gap_T-benchmarks}
\end{figure}

To demonstrate the robustness of DQIG compared to its variants, Figure \ref{fig:boxplot_time_Gap_T-benchmarks} presents boxplots of RPD (\%) for each instance set of Taillard dataset at scale $t = 120$. As depicted, DQIG achieves solutions with lower median and mean RPD values, along with lower standard deviations across almost all instance sets. This lower variability indicates DQIG's robustness in solving problem instances. 

When comparing the overall performance of DQIG with a dynamic portfolio to SQIG with a static portfolio, we see that DQIG achieves better solutions under the same computational time. This is because the Q-learning-based AOS mechanism favors operators with higher performance (higher Q-values), which leads to these operators being selected over and over, even if they cannot improve the solution (i.e., cycling over inefficient operators). This leads SQIG to a premature convergence. However, when embedding the dynamic portfolio determination into the Q-leaning-based AOS mechanism, the tabu mechanism helps to avoid cycling over inefficient operators by temporarily removing these non-improving operators from the portfolio. This enhances the chance to explore other operators, leading to discovering solutions that might remain unexplored.

According to Table \ref{tab: QIG vs. Benchmarks scale 60 Taillard} and Figure \ref{fig:boxplot_time_Gap_T-benchmarks}, we conclude that DQIG not only achieves solutions of higher quality compared to its variants but also demonstrates a more stable performance, particularly in handling larger instance sets (``tai\_200\_20'' and ``tai\_500\_20''). To support this conclusion and further investigate the performance of DQIG over larger and more complex instances, we extend our computational experiments to a larger and more challenging PFSP dataset, \textit{VRF-hard-large} dataset. Accordingly, Table \ref{tab: QIG vs. Benchmarks scale 60, 90, 120 VL} shows the performance of DQIG against SQIG over \textit{VRF-hard-large} dataset for different time scales of $ t \in \{60, 90, 120\}$. Table \ref{tab: QIG vs. Benchmarks scale 60, 90, 120 VL} indicates that DQIG significantly outperforms SQIG in solving complex instances. This outperformance highlights the contribution of dynamically adjusting the portfolio of operators versus a static portfolio, which is even more prominent for larger instances. As problem instances become larger, the combinatorial complexity and potential for suboptimal exploration increase. Embedding the tabu mechanism into the Q-learning-based AOS helps address these challenges by guiding exploration and avoiding local traps more effectively. This guides the search mechanism to converge to better solutions under the same computational time (i.e., faster convergence). 

{
\fontsize{6}{7}\selectfont
\setlength{\tabcolsep}{0.5pt}

\begin{tabularx}{1.55\textwidth}{
                       >{\hsize=0.03\linewidth}X
                       >{\hsize=0.03\linewidth}X 
                       >{\hsize=0.045\linewidth}X 
                       >{\hsize=0.045\linewidth}X
                       >{\hsize=0.045\linewidth}X 
                       >{\hsize=0.045\linewidth}X
                       >{\hsize=0.045\linewidth}X 
                       >{\hsize=0.045\linewidth}X
                       >{\hsize=0.045\linewidth}X 
                       >{\hsize=0.045\linewidth}X
                       >{\hsize=0.045\linewidth}X 
                       >{\hsize=0.045\linewidth}X
                       >{\hsize=0.045\linewidth}X 
                       >{\hsize=0.045\linewidth}X}

  \caption{Performance comparison of DQIG against its variant for \textit{VRF-hard-large} dataset for $t=60, 90, 120$}
  \label{tab: QIG vs. Benchmarks scale 60, 90, 120 VL} \\
  
  \toprule
  \multicolumn{2}{l}{Instance}	&	\multicolumn{12}{l}{Algorithms}	\\
  \hline
  && \multicolumn{4}{l}{$t=60$} & \multicolumn{4}{l}{$t=90$} & \multicolumn{4}{l}{$t=120$} \\
  \hline
  &&	\multicolumn{2}{l}{$\text{SQIG}$}	& \multicolumn{2}{l}{$\text{DQIG}$}	& \multicolumn{2}{l}{$\text{SQIG}$}	& \multicolumn{2}{l}{$\text{DQIG}$}	&	\multicolumn{2}{l}{$\text{SQIG}$}	& \multicolumn{2}{l}{$\text{DQIG}$}	\\ 
  \hline
  $n$ & $m$&	Average	&	Best	&	Average	&	Best &	Average	&	Best	&	Average	&	Best	&	Average	&	Best	&	Average	&	Best\\

  \midrule
  \endfirsthead
  \multicolumn{14}{c}{\tablename~\thetable~ (continued)} \\
  \toprule
  \multicolumn{2}{l}{Instance}	&	\multicolumn{12}{l}{Algorithms}	\\
  \hline
  &&	\multicolumn{2}{l}{$\text{IG}_\text{RS}$}	& \multicolumn{2}{l}{$\text{IG}_\text{PTL}$}	& \multicolumn{2}{l}{$\text{IG}_\text{FF}$}	& \multicolumn{2}{l}{$\text{IG}_\text{DPS}$}	&	\multicolumn{2}{l}{RIG}	&	\multicolumn{2}{l}{QIG}\\ 
  \hline
  $n$ & $m$&	Average	&	Best	&	Average	&	Best &	Average	&	Best	&	Average	&	Best	&	Average	&	Best	&	Average	&	Best\\
  
  \midrule
  \endhead
  \midrule
  \multicolumn{14}{r@{}}{\footnotesize To be continued ...}
  \endfoot
  \bottomrule
  \endlastfoot

100	&	20	&	0.669	&	0.604	&	\textbf{0.227}	&	0.126	&	0.558	&	0.461	&	\textbf{0.191}	&	0.099	&	0.480	&	0.394	&	\textbf{0.089}	&	0.027	\\
100	&	40	&	0.501	&	0.334	&	\textbf{0.230}	&	0.103	&	0.351	&	0.247	&	\textbf{0.019}	&	-0.080	&	0.182	&	0.081	&	\textbf{-0.069}	&	-0.140	\\
100	&	60	&	0.303	&	0.196	&	\textbf{0.221}	&	0.071	&	0.170	&	0.053	&	\textbf{0.065}	&	-0.037	&	0.072	&	0.006	&	\textbf{-0.042}	&	-0.125	\\
200	&	20	&	0.351	&	0.224	&	\textbf{0.149}	&	0.075	&	0.273	&	0.205	&	\textbf{0.059}	&	-0.019	&	0.219	&	0.150	&	\textbf{0.018}	&	-0.053	\\
200	&	40	&	0.359	&	0.259	&	\textbf{0.208}	&	0.111	&	0.207	&	0.132	&	\textbf{0.067}	&	-0.057	&	0.124	&	0.047	&	\textbf{-0.052}	&	-0.133	\\
200	&	60	&	0.358	&	0.240	&	\textbf{0.082}	&	-0.042	&	0.205	&	0.110	&	\textbf{-0.047}	&	-0.186	&	0.073	&	-0.023	&	\textbf{-0.134}	&	-0.262	\\
300	&	20	&	0.207	&	0.117	&	\textbf{0.021}	&	-0.074	&	0.162	&	0.072	&	\textbf{-0.031}	&	-0.102	&	0.125	&	0.028	&	\textbf{-0.064}	&	-0.127	\\
300	&	40	&	0.429	&	0.350	&	\textbf{0.155}	&	0.032	&	0.364	&	0.297	&	\textbf{0.059}	&	-0.042	&	0.251	&	0.170	&	\textbf{-0.007}	&	-0.115	\\
300	&	60	&	0.289	&	0.204	&	\textbf{0.321}	&	0.194	&	0.187	&	0.088	&	\textbf{0.196}	&	0.073	&	0.082	&	-0.012	&	\textbf{0.108}	&	-0.003	\\
400	&	20	&	0.165	&	0.109	&	\textbf{0.091}	&	0.059	&	0.113	&	0.052	&	\textbf{0.071}	&	0.035	&	0.079	&	0.026	&	\textbf{0.048}	&	0.010	\\
400	&	40	&	0.309	&	0.213	&	\textbf{0.204}	&	0.087	&	0.237	&	0.172	&	\textbf{0.117}	&	-0.020	&	0.175	&	0.123	&	\textbf{0.023}	&	-0.111	\\
400	&	60	&	0.150	&	0.055	&	\textbf{0.103}	&	0.006	&	0.068	&	-0.053	&	\textbf{-0.006}	&	-0.084	&	0.001	&	-0.093	&	\textbf{-0.075}	&	-0.155	\\
500	&	20	&	0.145	&	0.106	&	\textbf{0.049}	&	0.001	&	0.107	&	0.081	&	\textbf{0.026}	&	-0.015	&	0.093	&	0.065	&	\textbf{0.017}	&	-0.020	\\
500	&	40	&	0.223	&	0.126	&	\textbf{0.088}	&	0.012	&	0.173	&	0.074	&	\textbf{0.024}	&	-0.048	&	0.118	&	0.022	&	\textbf{-0.026}	&	-0.095	\\
500	&	60	&	0.333	&	0.224	&	\textbf{0.190}	&	0.122	&	0.252	&	0.151	&	\textbf{0.098}	&	0.053	&	0.184	&	0.089	&	\textbf{0.049}	&	0.004	\\
600	&	20	&	0.055	&	0.023	&	\textbf{0.008}	&	-0.019	&	0.035	&	-0.002	&	\textbf{-0.005}	&	-0.035	&	0.020	&	-0.011	&	\textbf{-0.014}	&	-0.049	\\
600	&	40	&	0.210	&	0.126	&	\textbf{0.166}	&	0.093	&	0.156	&	0.072	&	\textbf{0.125}	&	0.032	&	0.114	&	0.015	&	\textbf{0.073}	&	-0.016	\\
600	&	60	&	0.153	&	0.071	&	\textbf{0.087}	&	0.019	&	0.107	&	0.025	&	\textbf{0.027}	&	-0.057	&	0.063	&	-0.014	&	\textbf{-0.010}	&	-0.103	\\
700	&	20	&	0.101	&	0.072	&	\textbf{0.028}	&	-0.004	&	0.091	&	0.063	&	\textbf{0.011}	&	-0.014	&	0.074	&	0.044	&	\textbf{-0.013}	&	-0.034	\\
700	&	40	&	0.059	&	-0.016	&	\textbf{-0.023}	&	-0.099	&	0.028	&	-0.038	&	\textbf{-0.074}	&	-0.143	&	0.006	&	-0.050	&	\textbf{-0.105}	&	-0.181	\\
700	&	60	&	0.218	&	0.177	&	\textbf{0.058}	&	-0.004	&	0.175	&	0.111	&	\textbf{0.021}	&	-0.033	&	0.108	&	0.027	&	\textbf{-0.019}	&	-0.077	\\
800	&	20	&	0.182	&	0.162	&	\textbf{0.057}	&	0.009	&	0.171	&	0.152	&	\textbf{0.035}	&	-0.009	&	0.058	&	0.031	&	\textbf{0.022}	&	-0.021	\\
800	&	40	&	0.051	&	-0.005	&	\textbf{0.012}	&	-0.063	&	0.004	&	-0.048	&	\textbf{-0.038}	&	-0.110	&	-0.029	&	-0.081	&	\textbf{-0.081}	&	-0.153	\\
800	&	60	&	0.206	&	0.136	&	\textbf{0.120}	&	0.046	&	0.162	&	0.096	&	\textbf{0.093}	&	0.013	&	0.142	&	0.079	&	\textbf{0.054}	&	-0.021	\\

\\

\multicolumn{2}{l}{Average} &	0.251	&	0.171	&	\textbf{0.119}	&	0.036	&	0.181	&	0.107	&	\textbf{0.046}	&	-0.033	&	0.117	&	0.046	&	\textbf{-0.009}	&	-0.081	\\

\end{tabularx}}

\subsection{Comparison of DQIG with state-of-the-art IG algorithms} \label{sec: phase 2}

In this section, we compare the performance of DQIG with the state-of-the-art algorithms for PFSP. This analysis is done in detail per instance set (Subsection \ref{sec: phase 2 optimality gap detailed}) and overall on all datasets (Subsection \ref{sec: phase 2 optimality gap overall}). 

\subsubsection{Detailed instance-based competitiveness} \label{sec: phase 2 optimality gap detailed}

Table \ref{tab: QIG vs. SOTA scale 60 Taillard} shows the performance of DQIG and the state-of-the-art algorithms on \textit{Taillard} dataset under the same computational time. This table shows that DQIG beats all the state-of-the-art algorithms in terms of both ``Average'' and ``Best'' measures. 
The main difference between DQIG and the state-of-the-art algorithms for PFSP lies in the fact that the state-of-the-art algorithms commonly employ only a single perturbation operator, eliminating the need for an operator management mechanism. Therefore, the outperformance of DQIG can be attributed to its high exploration ability obtained by adding multiple perturbation operators and adaptive management of these operators using tabu and Q-learning-based AOS mechanisms. The results of the statistical test also confirm the statistical significance of DQIG against each of the other algorithms with 95\% confidence interval. For certain instance sets, such as ``tai\_20\_20'' and ``tai\_100\_5'', even though DQIG obtains better results, the difference between DQIG and other algorithms is not statistically significant since all algorithms can obtain solutions with minimal optimality gaps. 

\begin{figure}[H]
    \centering
    \includegraphics[scale = 0.35]{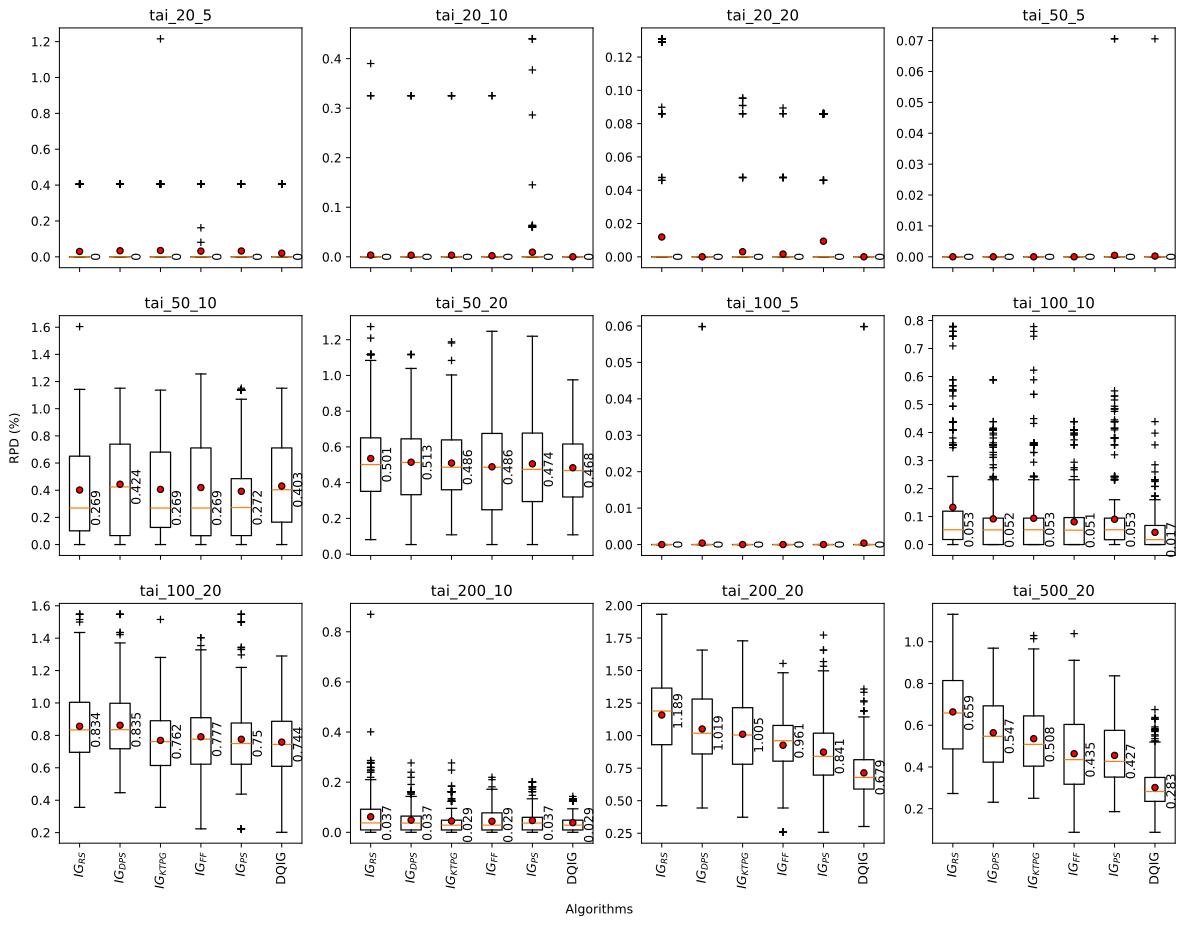}
    \caption{Boxplot of DQIG and the state-of-the-art algorithms based on RPD (\%) for scale $t=120$ for each instance set of \textit{Taillard} dataset} 
    \label{fig:boxplot_time_Gap_T_SOTA}
\end{figure}

To demonstrate the robustness of DQIG compared to the state-of-the-art algorithms, Figure \ref{fig:boxplot_time_Gap_T_SOTA} depicts the boxplots of RPD (\%) for \textit{Taillard} dataset at scale $t = 120$. As can be seen, for the majority of the instance sets, DQIG achieves a set of solutions with lower median and mean values as well as lower standard deviations. This implies that DQIG exhibits a more robust behavior than other algorithms, particularly when solving larger instance sets.

{
\fontsize{6}{7}\selectfont
\setlength{\tabcolsep}{0.5pt}

\begin{tabularx}{1.55\textwidth}{
                       >{\hsize=0.03\linewidth}X
                       >{\hsize=0.03\linewidth}X 
                       >{\hsize=0.045\linewidth}X 
                       >{\hsize=0.045\linewidth}X
                       >{\hsize=0.045\linewidth}X 
                       >{\hsize=0.045\linewidth}X
                       >{\hsize=0.045\linewidth}X 
                       >{\hsize=0.045\linewidth}X
                       >{\hsize=0.045\linewidth}X 
                       >{\hsize=0.045\linewidth}X
                       >{\hsize=0.045\linewidth}X 
                       >{\hsize=0.045\linewidth}X
                       >{\hsize=0.045\linewidth}X 
                       >{\hsize=0.045\linewidth}X}

  \caption{Performance comparison of DQIG against the state-of-the-art algorithms on \textit{Taillard} dataset for $t=60, 90, 120$}

  \label{tab: QIG vs. SOTA scale 60 Taillard} \\
  
  \toprule
  \multicolumn{2}{l}{Instance}	&	\multicolumn{12}{l}{Algorithms ($t=60$)}	\\
  \hline
  &&	\multicolumn{2}{l}{$\text{IG}_\text{RS}$}	& \multicolumn{2}{l}{$\text{IG}_\text{DPS}$}	& \multicolumn{2}{l}{$\text{IG}_\text{KTPG}$} &	\multicolumn{2}{l}{$\text{IG}_\text{FF}$}	& \multicolumn{2}{l}{$\text{IG}_\text{PS}$}		&	\multicolumn{2}{l}{DQIG}\\  
  
  \hline
  $n$ & $m$&	Average	&	Best	&	Average	&	Best &	Average	&	Best	&	Average	&	Best	&	Average	&	Best	&	Average	&	Best\\

  \midrule
  \endfirsthead
  \multicolumn{14}{c}{\tablename~\thetable~ (continued)} \\
  \toprule
  \multicolumn{2}{l}{Instance}	&	\multicolumn{12}{l}{Algorithms}	\\
  \hline
  &&	\multicolumn{2}{l}{$\text{IG}_\text{RS}$}	& \multicolumn{2}{l}{$\text{IG}_\text{DPS}$}	& \multicolumn{2}{l}{$\text{IG}_\text{KTPG}$} &	\multicolumn{2}{l}{$\text{IG}_\text{FF}$}	& \multicolumn{2}{l}{$\text{IG}_\text{PS}$}		&	\multicolumn{2}{l}{DQIG}\\  
  \hline
  $n$ & $m$&	Average	&	Best	&	Average	&	Best &	Average	&	Best	&	Average	&	Best	&	Average	&	Best	&	Average	&	Best\\

  \midrule
  \endhead
  \midrule
  \multicolumn{14}{r@{}}{\footnotesize To be continued ...}
  \endfoot
  \bottomrule
  \endlastfoot

20	&	5	&	0.036	&	0	&	0.036	&	0	&	0.040	&	0	&	0.034	&	0	&	0.037	&	0	&	\textbf{0.020}	&	0	\\
20	&	10	&	0.011	&	0	&	0.010	&	0	&	0.018	&	0	&	0.009	&	0	&	0.016	&	0	&	\textbf{0}	&	0	\\
20	&	20	&	0.023	&	0	&	\textbf{0.003}	&	0	&	0.008	&	0	&	\textbf{0.005}	&	0	&	0.014	&	0.013	&	\textbf{0.002}	&	0	\\
50	&	5	&	0.002	&	0	&	0.002	&	0	&	0.002	&	0	&	0.002	&	0	&	0.002	&	0	&	0.001	&	0	\\
50	&	10	&	0.470	&	0.290	&	0.505	&	0.310	&	0.482	&	0.290	&	0.475	&	0.310	&	0.448	&	0.279	&	0.494	&	0.307	\\
50	&	20	&	0.667	&	0.294	&	\textbf{0.617}	&	0.317	&	\textbf{0.612}	&	0.310	&	\textbf{0.593}	&	0.295	&	\textbf{0.605}	&	0.305	&	\textbf{0.611}	&	0.278	\\
100	&	5	&	0.009	&	0.008	&	0.010	&	0.008	&	0.010	&	0.008	&	0.008	&	0.008	&	0.010	&	0.008	&	0.008	&	0.008	\\
100	&	10	&	0.164	&	0.025	&	0.138	&	0.034	&	0.129	&	0.042	&	0.122	&	0.034	&	0.115	&	0.026	&	\textbf{0.091}	&	0.021	\\
100	&	20	&	0.992	&	0.554	&	0.996	&	0.632	&	0.894	&	0.573	&	0.883	&	0.593	&	0.860	&	0.608	&	0.897	&	0.569	\\
200	&	10	&	0.082	&	0.033	&	0.058	&	0.034	&	0.057	&	0.032	&	0.055	&	0.033	&	0.056	&	0.030	&	\textbf{0.049}	&	0.030	\\
200	&	20	&	1.249	&	0.880	&	1.144	&	0.819	&	1.084	&	0.766	&	0.978	&	0.728	&	0.943	&	0.729	&	\textbf{0.839}	&	0.576	\\
500	&	20	&	0.693	&	0.514	&	0.595	&	0.428	&	0.551	&	0.397	&	0.496	&	0.364	&	0.499	&	0.375	&	\textbf{0.338}	&	0.223	\\

\\

\multicolumn{2}{l}{Average} & 0.367	&	0.216	&	0.343	&	0.215	&	0.324	&	0.201	&	0.305	&	0.197	&	0.300	&	0.198	&	\textbf{0.279}	&	0.168	\\

	\\

\hline
\\

\multicolumn{2}{l}{Instance}	&	\multicolumn{12}{l}{Algorithms ($t=90$)}	\\
  \hline
  &&	\multicolumn{2}{l}{$\text{IG}_\text{RS}$}	& \multicolumn{2}{l}{$\text{IG}_\text{DPS}$}	& \multicolumn{2}{l}{$\text{IG}_\text{KTPG}$} &	\multicolumn{2}{l}{$\text{IG}_\text{FF}$}	& \multicolumn{2}{l}{$\text{IG}_\text{PS}$}		&	\multicolumn{2}{l}{DQIG}\\  
  
  \hline
  $n$ & $m$&	Average	&	Best	&	Average	&	Best &	Average	&	Best	&	Average	&	Best	&	Average	&	Best	&	Average	&	Best\\
\hline

\\

20	&	5	&	0.032	&	0	&	0.034	&	0	&	0.036	&	0	&	0.033	&	0	&	0.034	&	0	&	\textbf{0.020}	&	0	\\
20	&	10	&	\textbf{0.008}	&	0	&	0.005	&	0	&	\textbf{0.008}	&	0	&	\textbf{0.005}	&	0	&	\textbf{0.011}	&	0	&	\textbf{0}	&	0	\\
20	&	20	&	\textbf{0.016}	&	0	&	0.001	&	0	&	\textbf{0.006}	&	0	&	\textbf{0.003}	&	0	&	\textbf{0.010}	&	0	&	\textbf{0}	&	0	\\
50	&	5	&	0	&	0	&	0	&	0	&	0.000	&	0	&	0.000	&	0	&	0.001	&	0	&	0	&	0	\\
50	&	10	&	0.429	&	0.273	&	0.464	&	0.289	&	0.431	&	0.276	&	0.439	&	0.282	&	0.409	&	0.276	&	0.456	&	0.293	\\
50	&	20	&	0.589	&	0.264	&	\textbf{0.562}	&	0.242	&	\textbf{0.543}	&	0.250	&	\textbf{0.534}	&	0.244	&	\textbf{0.544}	&	0.259	&	\textbf{0.540}	&	0.240	\\
100	&	5	&	0.008	&	0.008	&	0.009	&	0.008	&	0.008	&	0.008	&	0.008	&	0.008	&	0.008	&	0.008	&	0.008	&	0.008	\\
100	&	10	&	0.145	&	0.025	&	0.108	&	0.027	&	0.111	&	0.025	&	0.097	&	0.023	&	0.100	&	0.025	&	\textbf{0.060}	&	0.021	\\
100	&	20	&	0.910	&	0.508	&	0.914	&	0.553	&	\textbf{0.800}	&	0.500	&	\textbf{0.820}	&	0.530	&	\textbf{0.811}	&	0.535	&	\textbf{0.812}	&	0.476	\\
200	&	10	&	0.072	&	0.033	&	0.052	&	0.033	&	0.051	&	0.031	&	0.052	&	0.033	&	0.054	&	0.029	&	\textbf{0.042}	&	0.036	\\
200	&	20	&	1.194	&	0.847	&	1.094	&	0.792	&	1.044	&	0.746	&	0.954	&	0.706	&	0.898	&	0.687	&	\textbf{0.767}	&	0.528	\\
500	&	20	&	0.677	&	0.509	&	0.578	&	0.411	&	0.542	&	0.396	&	0.476	&	0.357	&	0.473	&	0.342	&	\textbf{0.316}	&	0.211	\\

\\

\multicolumn{2}{l}{Average} & 0.340	&	0.206	&	0.318	&	0.196	&	0.299	&	0.186	&	0.285	&	0.182	&	0.279	&	0.181	&	\textbf{0.252}	&	0.151	\\

\hline
\\

\multicolumn{2}{l}{Instance}	&	\multicolumn{12}{l}{Algorithms ($t=120$)}	\\
  \hline
  &&	\multicolumn{2}{l}{$\text{IG}_\text{RS}$}	& \multicolumn{2}{l}{$\text{IG}_\text{DPS}$}	& \multicolumn{2}{l}{$\text{IG}_\text{KTPG}$} &	\multicolumn{2}{l}{$\text{IG}_\text{FF}$}	& \multicolumn{2}{l}{$\text{IG}_\text{PS}$}		&	\multicolumn{2}{l}{DQIG}\\  
  
  \hline
  $n$ & $m$&	Average	&	Best	&	Average	&	Best &	Average	&	Best	&	Average	&	Best	&	Average	&	Best	&	Average	&	Best\\
\hline

\\

20	&	5	&	\textbf{0.030}	&	0	&	0.034	&	0	&	0.035	&	0	&	0.032	&	0	&	0.032	&	0	&	\textbf{0.020}	&	0	\\
20	&	10	&	\textbf{0.003}	&	0	&	\textbf{0.003}	&	0	&	\textbf{0.003}	&	0	&	\textbf{0.002}	&	0	&	0.009	&	0	&	\textbf{0}	&	0	\\
20	&	20	&	0.012	&	0	&	0	&	0	&	0.003	&	0	&	0.002	&	0	&	0.009	&	0.009	&	0	&	0	\\
50	&	5	&	0	&	0	&	0	&	0	&	0	&	0	&	0	&	0	&	0	&	0	&	0	&	0	\\
50	&	10	&	0.401	&	0.269	&	0.444	&	0.276	&	0.406	&	0.273	&	0.420	&	0.282	&	0.392	&	0.276	&	0.431	&	0.293	\\
50	&	20	&	0.535	&	0.240	&	0.514	&	0.217	&	\textbf{0.510}	&	0.237	&	\textbf{0.488}	&	0.215	&	\textbf{0.505}	&	0.216	&	\textbf{0.483}	&	0.202	\\
100	&	5	&	0.008	&	0.008	&	0.008	&	0.008	&	0.008	&	0.008	&	0.008	&	0.008	&	0.008	&	0.008	&	0.008	&	0.008	\\
100	&	10	&	0.133	&	0.025	&	0.092	&	0.027	&	0.094	&	0.021	&	0.081	&	0.019	&	0.090	&	0.025	&	\textbf{0.043}	&	0.019	\\
100	&	20	&	0.856	&	0.486	&	0.863	&	0.553	&	\textbf{0.769}	&	0.478	&	0.791	&	0.521	&	\textbf{0.776}	&	0.502	&	\textbf{0.759}	&	0.436	\\
200	&	10	&	0.062	&	0.033	&	0.049	&	0.032	&	0.045	&	0.030	&	0.044	&	0.033	&	0.047	&	0.029	&	\textbf{0.038}	&	0.036	\\
200	&	20	&	1.159	&	0.790	&	1.051	&	0.764	&	1.011	&	0.723	&	0.926	&	0.691	&	0.874	&	0.660	&	\textbf{0.714}	&	0.468	\\
500	&	20	&	0.664	&	0.493	&	0.564	&	0.396	&	0.535	&	0.394	&	0.463	&	0.342	&	0.455	&	0.328	&	\textbf{0.301}	&	0.204	\\

\\

\multicolumn{2}{l}{Average} & 0.322	&	0.195	&	0.302	&	0.189	&	0.285	&	0.180	&	0.271	&	0.176	&	0.267	&	0.171	&	\textbf{0.233}	&	0.139	\\

\end{tabularx}}

To further compare the competitiveness of DQIG with the state-of-the-art over larger and more complex instances, we conduct similar comparison analyses on \textit{VRF-hard-large} dataset, presented in Tables \ref{tab: QIG vs. SOTA scale 60 VL} to \ref{tab: QIG vs. SOTA scale 120 VL} for time scales of $ t \in \{60, 90, 120\}$. Comparing the results of DQIG versus the state-of-the-art algorithms across different time scales $t=60, 90, 120$, we see that DQIG exhibits a faster convergence to better solutions from the beginning, and this superior performance is consistently maintained throughout the search. Furthermore, looking at ``Best'' columns in Table \ref{tab: QIG vs. SOTA scale 120 VL} with scale $t = 120$, we see that DQIG significantly improves the best-known solutions of the \textit{VRF-hard-large} dataset \citep{vallada2015new} by achieving negative RPDs.

{
\fontsize{6}{7}\selectfont
\setlength{\tabcolsep}{0.5pt}

\begin{tabularx}{1.55\textwidth}{
                       >{\hsize=0.03\linewidth}X
                       >{\hsize=0.03\linewidth}X 
                       >{\hsize=0.045\linewidth}X 
                       >{\hsize=0.045\linewidth}X
                       >{\hsize=0.045\linewidth}X 
                       >{\hsize=0.045\linewidth}X
                       >{\hsize=0.045\linewidth}X 
                       >{\hsize=0.045\linewidth}X
                       >{\hsize=0.045\linewidth}X 
                       >{\hsize=0.045\linewidth}X
                       >{\hsize=0.045\linewidth}X 
                       >{\hsize=0.045\linewidth}X
                       >{\hsize=0.045\linewidth}X 
                       >{\hsize=0.045\linewidth}X}

  \caption{Performance comparison of DQIG against the state-of-the-art algorithms on \textit{VRF-hard-large} dataset for $t=60$}
  \label{tab: QIG vs. SOTA scale 60 VL} \\
  
  \toprule
  \multicolumn{2}{l}{Instance}	&	\multicolumn{12}{l}{Algorithms}	\\
  \hline
  &&	\multicolumn{2}{l}{$\text{IG}_\text{RS}$}	& \multicolumn{2}{l}{$\text{IG}_\text{DPS}$}	& \multicolumn{2}{l}{$\text{IG}_\text{KTPG}$}	& \multicolumn{2}{l}{$\text{IG}_\text{FF}$}	&	\multicolumn{2}{l}{$\text{IG}_\text{PS}$}	&	\multicolumn{2}{l}{DQIG}\\ 
  \hline
  $n$ & $m$&	Average	&	Best	&	Average	&	Best &	Average	&	Best	&	Average	&	Best	&	Average	&	Best	&	Average	&	Best\\

  \midrule
  \endfirsthead
  \multicolumn{14}{c}{\tablename~\thetable~ (continued)} \\
  \toprule
  \multicolumn{2}{l}{Instance}	&	\multicolumn{12}{l}{Algorithms}	\\
  \hline
  &&	\multicolumn{2}{l}{$\text{IG}_\text{RS}$}	& \multicolumn{2}{l}{$\text{IG}_\text{DPS}$}	& \multicolumn{2}{l}{$\text{IG}_\text{KTPG}$}	& \multicolumn{2}{l}{$\text{IG}_\text{FF}$}	&	\multicolumn{2}{l}{$\text{IG}_\text{PS}$}	&	\multicolumn{2}{l}{DQIG}\\ 
  \hline
  $n$ & $m$&	Average	&	Best	&	Average	&	Best &	Average	&	Best	&	Average	&	Best	&	Average	&	Best	&	Average	&	Best\\

  \midrule
  \endhead
  \midrule
  \multicolumn{14}{r@{}}{\footnotesize To be continued ...}
  \endfoot
  \bottomrule
  \endlastfoot

100	&	20	&	0.606	&	0.162	&	0.579	&	0.248	&	0.485	&	0.162	&	0.469	&	0.089	&	0.460	&	0.106	&	\textbf{0.227}	&	0.126	\\
100	&	40	&	0.604	&	0.164	&	0.367	&	0.055	&	0.305	&	0.023	&	0.326	&	0.006	&	0.334	&	-0.006	&	\textbf{0.230}	&	0.103	\\
100	&	60	&	0.494	&	0.124	&	0.221	&	-0.051	&	0.211	&	-0.003	&	0.192	&	-0.094	&	0.208	&	-0.092	&	\textbf{0.221}	&	0.071	\\
200	&	20	&	0.800	&	0.507	&	0.702	&	0.471	&	0.552	&	0.281	&	0.519	&	0.221	&	0.513	&	0.170	&	\textbf{0.149}	&	0.075	\\
200	&	40	&	0.724	&	0.376	&	0.341	&	0.040	&	0.283	&	0.040	&	0.277	&	-0.024	&	0.297	&	-0.056	&	\textbf{0.208}	&	0.111	\\
200	&	60	&	0.568	&	0.160	&	0.277	&	0.016	&	0.232	&	0.007	&	0.234	&	-0.047	&	0.247	&	-0.048	&	\textbf{0.082}	&	-0.042	\\
300	&	20	&	0.629	&	0.425	&	0.461	&	0.297	&	0.394	&	0.116	&	0.341	&	0.077	&	0.322	&	0.097	&	\textbf{0.021}	&	-0.074	\\
300	&	40	&	0.793	&	0.431	&	0.415	&	0.125	&	0.323	&	0.146	&	0.344	&	0.010	&	0.295	&	0.034	&	\textbf{0.155}	&	0.032	\\
300	&	60	&	0.722	&	0.402	&	0.276	&	0.046	&	0.228	&	0.017	&	0.251	&	-0.004	&	0.238	&	0.001	&	\textbf{0.321}	&	0.194	\\
400	&	20	&	0.524	&	0.340	&	0.419	&	0.277	&	0.341	&	0.177	&	0.291	&	0.106	&	0.268	&	0.109	&	\textbf{0.091}	&	0.059	\\
400	&	40	&	0.763	&	0.455	&	0.355	&	0.157	&	0.232	&	0.063	&	0.258	&	0.047	&	0.238	&	-0.002	&	\textbf{0.204}	&	0.087	\\
400	&	60	&	0.747	&	0.439	&	0.200	&	-0.084	&	0.170	&	-0.072	&	0.160	&	-0.081	&	0.171	&	-0.110	&	\textbf{0.103}	&	0.006	\\
500	&	20	&	0.446	&	0.270	&	0.336	&	0.214	&	0.275	&	0.147	&	0.230	&	0.104	&	0.224	&	0.063	&	\textbf{0.049}	&	0.001	\\
500	&	40	&	0.661	&	0.391	&	0.271	&	0.074	&	0.199	&	-0.075	&	0.131	&	-0.115	&	0.144	&	-0.106	&	\textbf{0.088}	&	0.012	\\
500	&	60	&	0.759	&	0.450	&	0.211	&	-0.001	&	0.210	&	-0.025	&	0.202	&	-0.011	&	0.200	&	-0.009	&	\textbf{0.190}	&	0.122	\\
600	&	20	&	0.348	&	0.222	&	0.222	&	0.119	&	0.152	&	0.047	&	0.139	&	0.056	&	0.135	&	0.017	&	\textbf{0.008}	&	-0.019	\\
600	&	40	&	0.640	&	0.382	&	0.269	&	0.067	&	0.143	&	-0.039	&	0.130	&	-0.078	&	0.120	&	-0.110	&	\textbf{0.166}	&	0.093	\\
600	&	60	&	0.647	&	0.382	&	0.094	&	-0.126	&	0.065	&	-0.141	&	0.050	&	-0.127	&	0.044	&	-0.148	&	\textbf{0.087}	&	0.019	\\
700	&	20	&	0.347	&	0.231	&	0.244	&	0.141	&	0.151	&	0.039	&	0.145	&	0.038	&	0.139	&	0.019	&	\textbf{0.028}	&	-0.004	\\
700	&	40	&	0.545	&	0.321	&	0.186	&	0.021	&	0.018	&	-0.114	&	-0.009	&	-0.201	&	-0.003	&	-0.181	&	\textbf{-0.023}	&	-0.099	\\
700	&	60	&	0.587	&	0.331	&	0.086	&	-0.084	&	0.069	&	-0.120	&	0.054	&	-0.144	&	0.036	&	-0.170	&	\textbf{0.058}	&	-0.004	\\
800	&	20	&	0.274	&	0.193	&	0.186	&	0.089	&	0.129	&	0.033	&	0.113	&	0.037	&	0.101	&	0.028	&	\textbf{0.057}	&	0.009	\\
800	&	40	&	0.521	&	0.332	&	0.180	&	0.028	&	0.058	&	-0.082	&	0.027	&	-0.164	&	0.014	&	-0.183	&	\textbf{0.012}	&	-0.063	\\
800	&	60	&	0.635	&	0.412	&	0.145	&	-0.003	&	0.126	&	-0.075	&	0.110	&	-0.072	&	0.105	&	-0.082	&	\textbf{0.120}	&	0.046	\\

\\
\multicolumn{2}{l}{Average} &	0.599	&	0.329	&	0.293	&	0.089	&	0.223	&	0.023	&	0.208	&	-0.015	&	0.202	&	-0.027	&	\textbf{0.119}	&	0.036 \\

\end{tabularx}}

{
\fontsize{6}{7}\selectfont
\setlength{\tabcolsep}{0.5pt}

\begin{tabularx}{1.55\textwidth}{
                       >{\hsize=0.03\linewidth}X
                       >{\hsize=0.03\linewidth}X 
                       >{\hsize=0.045\linewidth}X 
                       >{\hsize=0.045\linewidth}X
                       >{\hsize=0.045\linewidth}X 
                       >{\hsize=0.045\linewidth}X
                       >{\hsize=0.045\linewidth}X 
                       >{\hsize=0.045\linewidth}X
                       >{\hsize=0.045\linewidth}X 
                       >{\hsize=0.045\linewidth}X
                       >{\hsize=0.045\linewidth}X 
                       >{\hsize=0.045\linewidth}X
                       >{\hsize=0.045\linewidth}X 
                       >{\hsize=0.045\linewidth}X}

  \caption{Performance comparison of DQIG against the state-of-the-art algorithms on \textit{VRF-hard-large} dataset for $t=90$}
  \label{tab: QIG vs. SOTA scale 90 VL} \\
  
  \toprule
  \multicolumn{2}{l}{Instance}	&	\multicolumn{12}{l}{Algorithms}	\\
  \hline
  &&	\multicolumn{2}{l}{$\text{IG}_\text{RS}$}	& \multicolumn{2}{l}{$\text{IG}_\text{DPS}$}	& \multicolumn{2}{l}{$\text{IG}_\text{KTPG}$}	& \multicolumn{2}{l}{$\text{IG}_\text{FF}$}	&	\multicolumn{2}{l}{$\text{IG}_\text{PS}$}	&	\multicolumn{2}{l}{DQIG}\\ 
  \hline
  $n$ & $m$&	Average	&	Best	&	Average	&	Best &	Average	&	Best	&	Average	&	Best	&	Average	&	Best	&	Average	&	Best\\

  \midrule
  \endfirsthead
  \multicolumn{14}{c}{\tablename~\thetable~ (continued)} \\
  \toprule
  \multicolumn{2}{l}{Instance}	&	\multicolumn{12}{l}{Algorithms}	\\
  \hline
  &&	\multicolumn{2}{l}{$\text{IG}_\text{RS}$}	& \multicolumn{2}{l}{$\text{IG}_\text{DPS}$}	& \multicolumn{2}{l}{$\text{IG}_\text{KTPG}$}	& \multicolumn{2}{l}{$\text{IG}_\text{FF}$}	&	\multicolumn{2}{l}{$\text{IG}_\text{PS}$}	&	\multicolumn{2}{l}{DQIG}\\ 
  \hline
  $n$ & $m$&	Average	&	Best	&	Average	&	Best &	Average	&	Best	&	Average	&	Best	&	Average	&	Best	&	Average	&	Best\\

  \midrule
  \endhead
  \midrule
  \multicolumn{14}{r@{}}{\footnotesize To be continued ...}
  \endfoot
  \bottomrule
  \endlastfoot

100	&	20	&	0.507	&	0.103	&	0.481	&	0.171	&	0.382	&	0.045	&	0.357	&	0.025	&	0.349	&	-0.011	&	\textbf{0.191}	&	0.099	\\
100	&	40	&	0.516	&	0.085	&	0.256	&	-0.028	&	0.219	&	-0.072	&	0.202	&	-0.098	&	0.216	&	-0.102	&	\textbf{0.019}	&	-0.080	\\
100	&	60	&	0.392	&	0.062	&	0.142	&	-0.099	&	0.123	&	-0.118	&	0.100	&	-0.165	&	0.127	&	-0.149	&	\textbf{0.065}	&	-0.037	\\
200	&	20	&	0.744	&	0.455	&	0.645	&	0.436	&	0.507	&	0.264	&	0.462	&	0.164	&	0.455	&	0.142	&	\textbf{0.059}	&	-0.019	\\
200	&	40	&	0.621	&	0.309	&	0.236	&	-0.021	&	0.204	&	-0.021	&	0.164	&	-0.103	&	0.194	&	-0.118	&	\textbf{0.067}	&	-0.057	\\
200	&	60	&	0.460	&	0.069	&	0.149	&	-0.121	&	0.138	&	-0.137	&	0.130	&	-0.141	&	0.124	&	-0.188	&	\textbf{-0.047}	&	-0.186	\\
300	&	20	&	0.601	&	0.410	&	0.436	&	0.267	&	0.378	&	0.100	&	0.296	&	0.050	&	0.282	&	0.041	&	\textbf{-0.031}	&	-0.102	\\
300	&	40	&	0.717	&	0.382	&	0.334	&	0.092	&	0.256	&	0.068	&	0.276	&	-0.021	&	0.227	&	-0.030	&	\textbf{0.059}	&	-0.042	\\
300	&	60	&	0.644	&	0.323	&	0.179	&	-0.044	&	0.156	&	-0.073	&	0.157	&	-0.116	&	0.147	&	-0.088	&	\textbf{0.196}	&	0.073	\\
400	&	20	&	0.503	&	0.328	&	0.396	&	0.261	&	0.313	&	0.130	&	0.268	&	0.092	&	0.241	&	0.095	&	\textbf{0.071}	&	0.035	\\
400	&	40	&	0.712	&	0.392	&	0.290	&	0.080	&	0.168	&	-0.032	&	0.187	&	-0.012	&	0.177	&	-0.042	&	\textbf{0.117}	&	-0.020	\\
400	&	60	&	0.678	&	0.380	&	0.121	&	-0.160	&	0.114	&	-0.163	&	0.095	&	-0.185	&	0.089	&	-0.182	&	\textbf{-0.006}	&	-0.084	\\
500	&	20	&	0.431	&	0.264	&	0.325	&	0.199	&	0.267	&	0.129	&	0.210	&	0.047	&	0.188	&	0.031	&	\textbf{0.026}	&	-0.015	\\
500	&	40	&	0.618	&	0.346	&	0.217	&	0.016	&	0.129	&	-0.111	&	0.103	&	-0.118	&	0.093	&	-0.130	&	\textbf{0.024}	&	-0.048	\\
500	&	60	&	0.702	&	0.392	&	0.163	&	-0.041	&	0.154	&	-0.032	&	0.142	&	-0.036	&	0.150	&	-0.033	&	\textbf{0.098}	&	0.053	\\
600	&	20	&	0.334	&	0.213	&	0.212	&	0.112	&	0.139	&	0.015	&	0.117	&	0.022	&	0.110	&	-0.015	&	\textbf{-0.005}	&	-0.035	\\
600	&	40	&	0.596	&	0.345	&	0.237	&	0.042	&	0.119	&	-0.109	&	0.082	&	-0.131	&	0.076	&	-0.174	&	\textbf{0.125}	&	0.032	\\
600	&	60	&	0.596	&	0.355	&	0.047	&	-0.149	&	0.008	&	-0.192	&	-0.013	&	-0.229	&	-0.005	&	-0.244	&	\textbf{0.027}	&	-0.057	\\
700	&	20	&	0.339	&	0.226	&	0.237	&	0.139	&	0.132	&	0.007	&	0.109	&	0.002	&	0.121	&	-0.003	&	\textbf{0.011}	&	-0.014	\\
700	&	40	&	0.518	&	0.293	&	0.159	&	-0.009	&	-0.030	&	-0.193	&	-0.068	&	-0.243	&	-0.048	&	-0.245	&	\textbf{-0.074}	&	-0.143	\\
700	&	60	&	0.551	&	0.285	&	0.051	&	-0.122	&	0.041	&	-0.161	&	0.016	&	-0.180	&	0.003	&	-0.206	&	\textbf{0.021}	&	-0.033	\\
800	&	20	&	0.263	&	0.184	&	0.179	&	0.083	&	0.104	&	0.010	&	0.095	&	-0.004	&	0.086	&	-0.019	&	\textbf{0.035}	&	-0.009	\\
800	&	40	&	0.488	&	0.305	&	0.155	&	0.010	&	0.029	&	-0.150	&	0.000	&	-0.184	&	-0.026	&	-0.246	&	\textbf{-0.038}	&	-0.110	\\
800	&	60	&	0.601	&	0.376	&	0.116	&	-0.062	&	0.091	&	-0.095	&	0.077	&	-0.098	&	0.069	&	-0.104	&	\textbf{0.093}	&	0.013	\\

\\
\multicolumn{2}{l}{Average} &	0.547	&	0.287	&	0.240	&	0.044	&	0.173	&	-0.037	&	0.149	&	-0.069	&	0.144	&	-0.084	&	\textbf{0.046}	&	-0.033	\\

\end{tabularx}}

{
\fontsize{6}{7}\selectfont
\setlength{\tabcolsep}{0.5pt}

\begin{tabularx}{1.55\textwidth}{
                       >{\hsize=0.03\linewidth}X
                       >{\hsize=0.03\linewidth}X 
                       >{\hsize=0.045\linewidth}X 
                       >{\hsize=0.045\linewidth}X
                       >{\hsize=0.045\linewidth}X 
                       >{\hsize=0.045\linewidth}X
                       >{\hsize=0.045\linewidth}X 
                       >{\hsize=0.045\linewidth}X
                       >{\hsize=0.045\linewidth}X 
                       >{\hsize=0.045\linewidth}X
                       >{\hsize=0.045\linewidth}X 
                       >{\hsize=0.045\linewidth}X
                       >{\hsize=0.045\linewidth}X 
                       >{\hsize=0.045\linewidth}X}

  \caption{Performance comparison of DQIG against the state-of-the-art algorithms on \textit{VRF-hard-large} dataset for $t=120$}
  \label{tab: QIG vs. SOTA scale 120 VL} \\
  
  \toprule
  \multicolumn{2}{l}{Instance}	&	\multicolumn{12}{l}{Algorithms}	\\
  \hline
  &&	\multicolumn{2}{l}{$\text{IG}_\text{RS}$}	& \multicolumn{2}{l}{$\text{IG}_\text{DPS}$}	& \multicolumn{2}{l}{$\text{IG}_\text{KTPG}$}	& \multicolumn{2}{l}{$\text{IG}_\text{FF}$}	&	\multicolumn{2}{l}{$\text{IG}_\text{PS}$}	&	\multicolumn{2}{l}{DQIG}\\ 
  \hline
  $n$ & $m$&	Average	&	Best	&	Average	&	Best &	Average	&	Best	&	Average	&	Best	&	Average	&	Best	&	Average	&	Best\\

  \midrule
  \endfirsthead
  \multicolumn{14}{c}{\tablename~\thetable~ (continued)} \\
  
  \toprule
  \multicolumn{2}{l}{Instance}	&	\multicolumn{12}{l}{Algorithms}	\\
  \hline
  &&	\multicolumn{2}{l}{$\text{IG}_\text{RS}$}	& \multicolumn{2}{l}{$\text{IG}_\text{DPS}$}	& \multicolumn{2}{l}{$\text{IG}_\text{KTPG}$}	& \multicolumn{2}{l}{$\text{IG}_\text{FF}$}	&	\multicolumn{2}{l}{$\text{IG}_\text{PS}$}	&	\multicolumn{2}{l}{DQIG}\\ 
  \hline
  $n$ & $m$&	Average	&	Best	&	Average	&	Best &	Average	&	Best	&	Average	&	Best	&	Average	&	Best	&	Average	&	Best\\

  \midrule
  \endhead
  \midrule
  \multicolumn{14}{r@{}}{\footnotesize To be continued ...}
  \endfoot
  \bottomrule
  \endlastfoot

100	&	20	&	0.428	&	0.027	&	0.412	&	0.139	&	0.309	&	-0.034	&	0.284	&	-0.056	&	0.273	&	-0.052	&	\textbf{0.089}	&	0.027	\\
100	&	40	&	0.434	&	0.014	&	0.158	&	-0.127	&	0.135	&	-0.168	&	0.133	&	-0.174	&	0.121	&	-0.196	&	\textbf{-0.069}	&	-0.140	\\
100	&	60	&	0.324	&	0.036	&	0.062	&	-0.150	&	0.051	&	-0.154	&	0.031	&	-0.226	&	0.036	&	-0.181	&	\textbf{-0.042}	&	-0.125	\\
200	&	20	&	0.710	&	0.419	&	0.605	&	0.370	&	0.425	&	0.212	&	0.410	&	0.143	&	0.403	&	0.087	&	\textbf{0.018}	&	-0.053	\\
200	&	40	&	0.551	&	0.229	&	0.157	&	-0.088	&	0.146	&	-0.110	&	0.101	&	-0.184	&	0.120	&	-0.149	&	\textbf{-0.052}	&	-0.133	\\
200	&	60	&	0.377	&	0.005	&	0.069	&	-0.183	&	0.075	&	-0.193	&	0.054	&	-0.267	&	0.038	&	-0.220	&	\textbf{-0.134}	&	-0.262	\\
300	&	20	&	0.583	&	0.387	&	0.418	&	0.258	&	0.342	&	0.081	&	0.259	&	0.035	&	0.243	&	0.008	&	\textbf{-0.064}	&	-0.127	\\
300	&	40	&	0.676	&	0.349	&	0.286	&	0.024	&	0.213	&	0.002	&	0.220	&	-0.088	&	0.185	&	-0.081	&	\textbf{-0.007}	&	-0.115	\\
300	&	60	&	0.584	&	0.250	&	0.097	&	-0.123	&	0.088	&	-0.129	&	0.087	&	-0.235	&	0.071	&	-0.166	&	\textbf{0.108}	&	-0.003	\\
400	&	20	&	0.487	&	0.324	&	0.381	&	0.240	&	0.306	&	0.106	&	0.248	&	0.088	&	0.217	&	0.090	&	\textbf{0.048}	&	0.010	\\
400	&	40	&	0.671	&	0.353	&	0.253	&	0.044	&	0.123	&	-0.069	&	0.162	&	-0.044	&	0.143	&	-0.085	&	\textbf{0.023}	&	-0.111	\\
400	&	60	&	0.633	&	0.334	&	0.058	&	-0.197	&	0.060	&	-0.197	&	0.036	&	-0.238	&	0.025	&	-0.240	&	\textbf{-0.075}	&	-0.155	\\
500	&	20	&	0.421	&	0.257	&	0.319	&	0.196	&	0.252	&	0.080	&	0.189	&	0.028	&	0.174	&	0.020	&	\textbf{0.017}	&	-0.020	\\
500	&	40	&	0.583	&	0.279	&	0.178	&	-0.024	&	0.091	&	-0.139	&	0.079	&	-0.147	&	0.067	&	-0.160	&	\textbf{-0.026}	&	-0.095	\\
500	&	60	&	0.657	&	0.361	&	0.124	&	-0.086	&	0.108	&	-0.054	&	0.090	&	-0.078	&	0.097	&	-0.068	&	\textbf{0.049}	&	0.004	\\
600	&	20	&	0.324	&	0.212	&	0.205	&	0.104	&	0.120	&	-0.007	&	0.103	&	-0.002	&	0.093	&	-0.030	&	\textbf{-0.014}	&	-0.049	\\
600	&	40	&	0.563	&	0.317	&	0.209	&	0.020	&	0.091	&	-0.146	&	0.070	&	-0.149	&	0.056	&	-0.202	&	\textbf{0.073}	&	-0.016	\\
600	&	60	&	0.563	&	0.314	&	0.011	&	-0.164	&	-0.025	&	-0.223	&	-0.046	&	-0.262	&	-0.029	&	-0.264	&	\textbf{-0.010}	&	-0.103	\\
700	&	20	&	0.330	&	0.210	&	0.231	&	0.136	&	0.122	&	-0.007	&	0.094	&	-0.007	&	0.115	&	-0.007	&	\textbf{-0.013}	&	-0.034	\\
700	&	40	&	0.491	&	0.268	&	0.133	&	-0.031	&	-0.055	&	-0.260	&	-0.095	&	-0.276	&	-0.068	&	-0.280	&	\textbf{-0.105}	&	-0.181	\\
700	&	60	&	0.520	&	0.254	&	0.021	&	-0.157	&	-0.003	&	-0.205	&	-0.018	&	-0.207	&	-0.038	&	-0.225	&	\textbf{-0.019}	&	-0.077	\\
800	&	20	&	0.258	&	0.180	&	0.175	&	0.082	&	0.096	&	-0.002	&	0.087	&	-0.011	&	0.081	&	-0.023	&	\textbf{0.022}	&	-0.021	\\
800	&	40	&	0.464	&	0.287	&	0.137	&	-0.015	&	-0.004	&	-0.178	&	-0.019	&	-0.193	&	-0.043	&	-0.269	&	\textbf{-0.081}	&	-0.153	\\
800	&	60	&	0.580	&	0.362	&	0.090	&	-0.083	&	0.077	&	-0.105	&	0.049	&	-0.110	&	0.044	&	-0.117	&	\textbf{0.054}	&	-0.021	\\

\\
\multicolumn{2}{l}{Average} &	0.509	&	0.251	&	0.200	&	0.008	&	0.131	&	-0.079	&	0.109	&	-0.111	&	0.101	&	-0.117	&	\textbf{-0.009}	&	-0.081	\\

\end{tabularx}
}

\subsubsection{Overall competitiveness}
\label{sec: phase 2 optimality gap overall}

\begin{figure}[H]
    \centering
    \includegraphics[scale = 0.4]{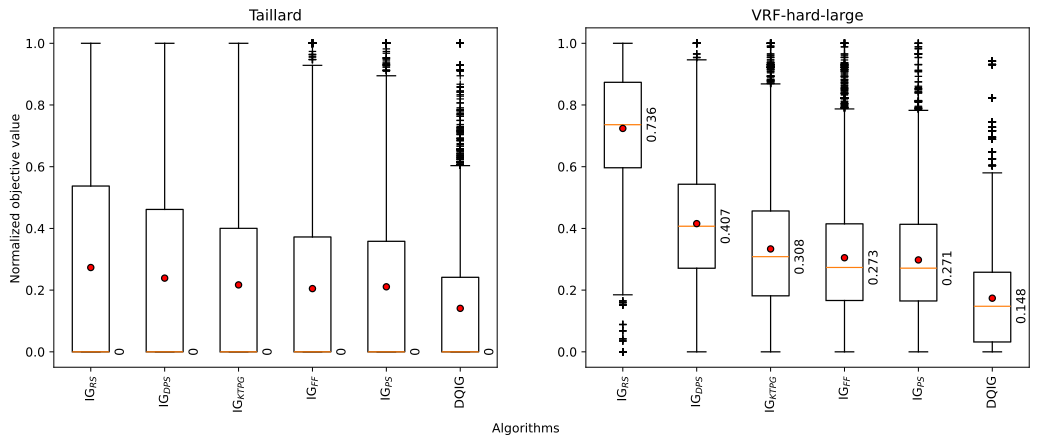}
    \caption{Boxplot of DQIG and the state-of-the-art algorithms based on normalized objective value over all instances of each dataset} 
    \label{fig:boxplot_Time_overall_normalized_SOTA}
\end{figure}

Table \ref{tab: overall_time_SOTA} together with Fig. \ref{fig:boxplot_Time_overall_normalized_SOTA} demonstrate the overall competitive performance of DQIG versus the state-of-the-art algorithms. Table \ref{tab: overall_time_SOTA} summarizes the performance of the algorithms for different time scales over the whole Taillard and \textit{VRF-hard-large} datasets. Figure \ref{fig:boxplot_Time_overall_normalized_SOTA} illustrates the robustness of algorithms over the whole two datasets. To depict the overall boxplots, we first need to normalize the objective function values for each instance set. For this aim, we normalize the corresponding values in a common range of $[0,1]$ using $x_N = (x - x_{min})/(x_{max} - x_{min})$, where $x_N$ and $x$ are the normalized and real values, respectively. In addition, $x_{min}$ and $x_{max}$ are the minimum and maximum values over all instances, respectively. This normalization is performed separately for each dataset. Looking at both datasets, DQIG yields better overall performance with lower median and mean values as well as lower standard deviations. 

Accordingly, at time scale $t=120$, DQIG improves the performance of the state-of-the-art algorithms $IG_{RS}$ \citep{ruiz2007simple} by 73\%, $IG_{DPS}$ \citep{dubois2017iterated} by 55\%, $IG_{KTPG}$ \citep{kizilay2019variable} by 46\%, $IG_{FF}$ \citep{fernandez2019best} by 41\%, and $IG_{PS}$ \citep{pagnozzi2019automatic} by 39\% on average. This overall analysis shows that the proposed operator management framework of this paper is the current state-of-the-art algorithm for solving the PFSP. 

The main drivers behind DQIG’s success lie in its effective operator management framework. Unlike conventional state-of-the-art algorithms that rely on a single perturbation operator, DQIG integrates multiple operators and employs an adaptive operator management strategy using tabu search and Q-learning-based AOS mechanisms. This multi-operator framework allows DQIG to dynamically balance exploration and exploitation throughout the search process, adapting to the search state more effectively than static or single-operator algorithms. The adaptability of DQIG is particularly advantageous when dealing with the diverse characteristics of PFSP instances. This allows DQIG’s to switch between different perturbation operators, enabling the algorithm to escape local optima and explore promising regions more thoroughly. Additionally, the use of Q-learning allows the algorithm to learn and prioritize the most effective operators based on the performance feedback, leading to improved solution quality and search efficiency over time.

{
\fontsize{5.5}{5.5}\selectfont
\setlength{\tabcolsep}{0.5pt}

\begin{tabularx}{1\textwidth}{
                       >{\hsize=0.12\linewidth}X
                       >{\hsize=0.06\linewidth}X 
                       >{\hsize=0.07\linewidth}X
                       >{\hsize=0.09\linewidth}X
                       >{\hsize=0.06\linewidth}X
                       >{\hsize=0.06\linewidth}X
                       >{\hsize=0.08\linewidth}X
                       >{\hsize=0.06\linewidth}X 
                       >{\hsize=0.07\linewidth}X
                       >{\hsize=0.09\linewidth}X
                       >{\hsize=0.06\linewidth}X
                       >{\hsize=0.06\linewidth}X
                       >{\hsize=0.08\linewidth}X 
                       >{\hsize=0.06\linewidth}X
                       >{\hsize=0.07\linewidth}X
                       >{\hsize=0.09\linewidth}X
                       >{\hsize=0.06\linewidth}X
                       >{\hsize=0.06\linewidth}X
                       >{\hsize=0.08\linewidth}X}

  \caption{Overall performance comparison of DQIG with state-of-the-art algorithms}
  \label{tab: overall_time_SOTA} \\
  
  \toprule
 Dataset & \multicolumn{6}{l}{$t=60$} & \multicolumn{6}{l}{$t=90$} & \multicolumn{6}{l}{$t=120$} \\
 \hline
   & $\text{IG}_\text{RS}$ & $\text{IG}_\text{DPS}$ & $\text{IG}_\text{KTPG}$ & $\text{IG}_\text{FF}$ & $\text{IG}_\text{PS}$ & DQIG & $\text{IG}_\text{RS}$ & $\text{IG}_\text{DPS}$ & $\text{IG}_\text{KTPG}$ & $\text{IG}_\text{FF}$ & $\text{IG}_\text{PS}$ & DQIG & $\text{IG}_\text{RS}$ & $\text{IG}_\text{DPS}$ & $\text{IG}_\text{KTPG}$ & $\text{IG}_\text{FF}$ & $\text{IG}_\text{PS}$ & DQIG   \\

  \midrule
  \endfirsthead
  \multicolumn{19}{c}{\tablename~\thetable~ (continued)} \\
  \toprule
 
 Dataset & \multicolumn{6}{l}{$t=60$} & \multicolumn{6}{l}{$t=90$} & \multicolumn{6}{l}{$t=120$} \\
 \hline
   & $\text{IG}_\text{RS}$ & $\text{IG}_\text{DPS}$ & $\text{IG}_\text{KTPG}$ & $\text{IG}_\text{FF}$ & $\text{IG}_\text{PS}$ & DQIG & $\text{IG}_\text{RS}$ & $\text{IG}_\text{DPS}$ & $\text{IG}_\text{KTPG}$ & $\text{IG}_\text{FF}$ & $\text{IG}_\text{PS}$ & DQIG & $\text{IG}_\text{RS}$ & $\text{IG}_\text{DPS}$ & $\text{IG}_\text{KTPG}$ & $\text{IG}_\text{FF}$ & $\text{IG}_\text{PS}$ & DQIG   \\
 
  \midrule
  \endhead
  \midrule
  \multicolumn{19}{r@{}}{\footnotesize To be continued ...}
  \endfoot
  \bottomrule
  \endlastfoot
Taillard	&	0.367	&	0.343	&	0.324	&	0.305	&	0.3	&	\textbf{0.279}	&	0.340	&	0.318	&	0.299	&	0.285	&	0.279	&	\textbf{0.252}	&	0.322	&	0.302	&	0.285	&	0.271	&	0.267	&	\textbf{0.233}	\\
VRF-large	&	0.599	&	0.293	&	0.223	&	0.208	&	0.202	&	\textbf{0.119}	&	0.547	&	0.240	&	0.173	&	0.149	&	0.144	&	\textbf{0.046}	&	0.509	&	0.200	&	0.131	&	0.109	&	0.101	&	\textbf{-0.009}	\\

\\
Average & 0.483	&	0.318	&	0.274	&	0.257	&	0.251	&	\textbf{0.199}	&	0.444	&	0.279	&	0.236	&	0.217	&	0.212	&	\textbf{0.149}	&	0.416	&	0.251	&	0.208	&	0.190	&	0.184	&	\textbf{0.112}	\\

\end{tabularx}}

\section{Conclusions} \label{sec: conclusion}

The performance of meta-heuristics in solving combinatorial optimization problems (COPs) can be enhanced by using multiple search operators with diverse characteristics, as the non-stationary nature of COP's search space benefits from a portfolio of operators, instead of a single operator. However, the challenge lies in determining which operators to include in the portfolio and how to effectively manage them during the search process. Static approaches that pre-determine operator portfolios based on the literature, expert knowledge, or pre-processing have limitations, including time-intensive pre-processing, poor adaptability to new instances, and the risk of suboptimal operator selection. Therefore, dynamic approaches are needed to adapt the portfolio and operator selection strategy based on the operator's performance and the state of the search.

In this paper, we propose a framework for dynamically managing search operators within meta-heuristics. This framework operates dynamically in two key phases: First, by adopting the idea of tabu search, it updates the set of active operators based on their short-term performance by temporarily excluding those operators that demonstrate unsuccessful performance in improving the solution. Subsequently, Q-learning is employed to adaptively select the most appropriate operator from the active set by taking into account the current state of the search, the historical performance of the operators, and the anticipated future gains.

To evaluate the proposed framework, we compare its performance to the state-of-the-art algorithms for the permutation flowshop scheduling problem (PFSP) on established benchmark datasets from the literature. First, we investigate whether incorporating the tabu search idea and Q-learning for managing the perturbation operators improves the performance of an iterated greedy (IG) algorithm. Second, the competitiveness of our framework has been tested against five state-of-the-art algorithms from the literature. The numerical results show the advantages of including both of the above-mentioned ideas for managing operators, as well as the superiority of the proposed framework in solving PFSP in comparison to state-of-the-art algorithms. The proposed DQIG framework improves the performance of the state-of-the-art algorithms $IG_{RS}$ \citep{ruiz2007simple} by 73\%, $IG_{DPS}$ \citep{dubois2017iterated} by 55\%, $IG_{KTPG}$ \citep{kizilay2019variable} by 46\%, $IG_{FF}$ \citep{fernandez2019best} by 41\%, and $IG_{PS}$ \citep{pagnozzi2019automatic} by 39\% on average.

Researchers and practitioners can achieve several benefits by using this framework. The dynamic adjustment of the operator set enhances the exploration-exploitation ability of the search process, ensuring that only the most promising operators are utilized. The integration of Q-learning facilitates a data-driven approach to operator selection, optimizing the decision-making process based on empirical performance metrics. Consequently, this framework improves the convergence speed and solution quality without adding significant computational overhead, making it a valuable tool for addressing complex optimization problems for which large sets of search operators exist.

This study opens up several opportunities for future research. First, through our comparative analysis, we show the effectiveness of the tabu mechanism within our proposed operator management framework. To further enhance the effectiveness of the operator management framework, we suggest enhancing the Q-learning-based Adaptive Operator Selection (AOS). Currently, the reward function in AOS considers the proportional improvement in the objective function. Future work could involve expanding this reward function to integrate additional factors, such as the operator's operation time and iteration number. In the early stages of the search, where improvements are generally easier to achieve, rewards could be down-weighted to encourage broader exploration. In contrast, as the search progresses and improvements become more difficult, increasing the rewards would prioritize operators that can make more refined gains. Second, the current tabu mechanism uniformly excludes operators based on their failure to improve the solution without differentiating between the beginning and end stages of the search. Our comparison of DQIG and SQIG demonstrates the effectiveness of this fixed exclusion mechanism. However, we expect that introducing an adaptive exclusion mechanism could further enhance the performance of the tabu mechanism. Specifically, in the early stages of the search, where exploring a wide range of solutions is critical, a more lenient exclusion mechanism could facilitate broader exploration. Conversely, a more restrictive exclusion mechanism could be employed as the search progresses and the focus shifts toward intensifying around high-quality solutions. An adaptive approach is expected to better balance exploration and exploitation, potentially leading to more efficient and effective search processes.

\appendix
\section{NEH heuristic \& insertion neighborhood local search} \label{apendix: NEH}

{\small
\vspace{0.1cm}
\SetAlgoCaptionSeparator{.}
\IncMargin{1em}

\begin{algorithm}[H]
\caption{Pseudo-code of the NEH heuristic}
\label{alg: NEH}

\SetKwFunction{FMain}{NEH}
\SetKwProg{Fn}{Function}{:}{}

\newcommand\mycommfont[1]{\footnotesize\ttfamily\textcolor{blue}{#1}}
\SetCommentSty{mycommfont}


\DontPrintSemicolon

\KwOut{$\Pi$ \tcp*[r]{A complete sequence of jobs}} 

\Fn{\FMain{}}{

    $\Pi := \emptyset$ \tcp*[r]{Start by an empty sequence $\Pi$}
    $\Pi^S := \texttt{sort()}$ \tcp*[r]{Sort jobs $j \in N$ in descending order based on $\sum_{i \in M} p_{ij}$}
    
    $\Pi := \Pi \cup \pi_1$ \tcp*[r]{$\pi_1$: the job in the first position of $\Pi^S$}
    $\Pi^S := \Pi^S \setminus \{\pi_1 \}$ \tcp*[r]{remove $\pi_1$ from $\Pi^S$}
    \While{$\Pi^S \neq \emptyset$}{
        $k^* := \argmin\limits_k C_{max} (\texttt{insert}(\Pi,\pi_1,k))$ \tcp*[r]{Find the best position $k^*$ in $\Pi$ to insert $\pi_1$} \tcp*[r]{$\texttt{insert}(\Pi,\pi_j,k)$: inserts $\pi_j$ in $k$th position of sequence $\Pi$ ($k \in K, |K| = |\Pi|+1$)}
        
        $\Pi := \texttt{insert}(\Pi, \pi_1, k^*)$\\ 
        $\Pi := \texttt{localSearch}(\Pi)$ \tcp*[r]{Optional: apply local search on partial sequence $\Pi$} 
        $\Pi^S := \Pi^S \setminus \{\pi_1 \}$\\
        }
    \Return $\Pi$
}

\end{algorithm}
\DecMargin{1em}
\vspace{0.1cm}
}

{\small
\vspace{0.1cm}
\SetAlgoCaptionSeparator{.}
\IncMargin{1em}

\begin{algorithm}[H]
\caption{Pseudo-code of insertion neighborhood \texttt{localSearch}}
\label{alg: insertLocalSearch}

\SetKwFunction{FMain}{localSearch}
\SetKwProg{Fn}{Function}{:}{}

\newcommand\mycommfont[1]{\footnotesize\ttfamily\textcolor{blue}{#1}}
\SetCommentSty{mycommfont}


\DontPrintSemicolon

\KwIn{$\Pi$ \tcp*[r]{A given solution}}
\KwOut{$\Pi$ \tcp*[r]{A local optimum}}
\Fn{\FMain{$\Pi$}}{
    $C^* := C_{max}(\Pi)$ \tcp*[r]{Remember $C_{max}$ of the best solution found during the local search}
    $\Pi^* := \Pi$ \tcp*[r]{Remember the best solution found during the local search}

    \While{\upshape \texttt{terminationCriterion()}}{

        $\Pi^S := \texttt{shuffle}(\Pi)$ \tcp*[r]{Shuffle sequence $\Pi$}
    \While{$\Pi^S \neq \emptyset$}{

        $k^* := \argmin\limits_k C_{max} (\texttt{insert}(\Pi,\pi_1,k))$\\
        $\Pi := \texttt{insert}(\Pi, \pi_1, k^*)$\\

        \If(\hspace{4.5cm}\tcp*[h]{Update best solution found during the local search}){\upshape \texttt{better}($\Pi, \Pi^*$)}
            {
                $C^* := C_{max}(\Pi)$ \\
                $\Pi^* := \Pi$
            }
        
        $\Pi^S := \Pi^S \setminus \{\pi_1 \}$\\
    }
        }
    \Return $\Pi^*$

}

\end{algorithm}
\DecMargin{1em}
\vspace{0.1cm}
}

\fontsize{8}{10}\selectfont
\setlength{\bibsep}{0pt plus 0.3ex}

\bibliography{mybibfile}

\end{document}